\documentclass[twocolumn,final,10pt]{elsarticle}
\usepackage{medima}
\usepackage[english]{babel} 

\usepackage[full]{textcomp}
\renewcommand{\copyright}{\textcopyright}

\usepackage{mathdesign}
\usepackage{comment}
\usepackage{multirow}
\usepackage{booktabs}
\usepackage{subcaption}
\usepackage[justification=centering]{caption}
\usepackage[dvipsnames]{xcolor}
\usepackage{pgfplots}
\usepackage{soul}
\usetikzlibrary{calc}

\usepackage[normalem]{ulem}

\usepackage{makecell, cellspace, caption}

\usepackage{amssymb} 
\usepackage{amsfonts}
\usepackage{amsmath}
\usepackage{amsthm}
\usepackage{amsmath,amsfonts,amsthm,bm} 
\usepackage{hyperref} 
\hypersetup{
        colorlinks
}
\usepackage{algorithm} 
\usepackage{algpseudocode} 

\usepackage[switch,displaymath]{lineno}

\usepackage{etoolbox}
\usepackage{graphicx}
\usepackage{times}
\usepackage{epsfig}
\usepackage{tabularx}
\usepackage{mathtools}
\usepackage{color, colortbl}
\usepackage{enumitem}
\usepackage{hhline}
\usepackage{bbm} 

\usepackage{amsmath,amsfonts,bm}

\def\eqref#1{equation~\ref{#1}}

\def\1{\bm{1}}

\def\rvl{{\mathbf{l}}}

\def\rvs{{\mathbf{s}}}

\def\vp{{\bm{p}}}

\def\vx{{\bm{x}}}
\def\vy{{\bm{y}}}

\DeclareMathAlphabet{\mathsfit}{\encodingdefault}{\sfdefault}{m}{sl}
\SetMathAlphabet{\mathsfit}{bold}{\encodingdefault}{\sfdefault}{bx}{n}

\def\sP{{\mathbb{P}}}

\def\sR{{\mathbb{R}}}

\newcommand{\ceq}{\stackrel{\mathclap{\normalfont\mbox{c}}}{=}}

\newcommand{\xx}{\mathbf{x}}

\newcommand{\yy}{\mathbf{y}}

\journal{Medical Image Analysis}

\title{Neighbor-Aware Calibration of Segmentation Networks with Penalty-Based Constraints} 

\algnewcommand\algorithmicinput{\textbf{Input:}}
\algnewcommand\Input{\item[\algorithmicinput]}
\algnewcommand\algorithmicoutput{\textbf{Output:}}
\algnewcommand\Output{\item[\algorithmicoutput]}

\newcommand{\ours}{\mbox{NACL}}

\verso{Murugesan \textit{et~al.}}

\author[1]{Balamurali \snm{Murugesan}\corref{cor1}}
\cortext[cor1]{Corresponding author: \url{balamurali.murugesan.1@ens.etsmtl.ca}}
\author[1]{Sukesh \snm{Adiga Vasudeva}}  
\author[2]{Bingyuan \snm{Liu}}

\author[1]{Herve \snm{Lombaert}}

\author[1]{Ismail \snm{Ben Ayed}}

\author[1]{Jose \snm{Dolz}}

\address[1]{ÉTS Montréal, Canada}
\address[2]{Amazon, Canada}

\received{1 September 2022}

\newcommand\und[1]{\underline{#1}}
\newcommand{\rev}[1]{{\color{black}#1}}

\begin{document}
\begin{frontmatter}
\begin{abstract}

Ensuring reliable confidence scores from 
deep neural networks is of paramount significance in critical decision-making systems, particularly in real-world domains such as healthcare. 
Recent literature on calibrating deep segmentation networks has resulted in substantial progress. Nevertheless, these approaches are strongly inspired by the advancements in classification tasks, and thus their uncertainty 
is usually modeled by leveraging the information of individual pixels, disregarding the local structure of the object of interest. Indeed, only the recent \textit{Spatially Varying Label Smoothing (SVLS)} approach considers pixel spatial relationships across classes, by softening the pixel label assignments with a discrete spatial Gaussian kernel. In this work, we first present a constrained optimization perspective of SVLS and demonstrate that it enforces an implicit constraint on soft class proportions of surrounding pixels. Furthermore, our analysis shows that SVLS lacks a mechanism to balance the contribution of the constraint with the primary objective, potentially hindering the optimization process. 
Based on these observations, 
we propose \rev{NACL (Neighbor Aware CaLibration)}, 
a principled and simple solution based on equality constraints on the logit values, which enables to control explicitly both the enforced constraint and the weight of the penalty, offering more flexibility. Comprehensive experiments on a wide variety of well-known segmentation benchmarks demonstrate the superior calibration performance of the proposed approach, without affecting its discriminative power. Furthermore, ablation studies empirically show the model agnostic nature of our approach, which can be used to train a wide span of deep segmentation networks. 
The code is available at \url{https://github.com/Bala93/MarginLoss}

\end{abstract}

\end{frontmatter}

\setlength{\parskip}{3pt}

\section{Introduction}
\label{sec:intro}
Despite the remarkable progress made by deep neural networks (DNNs) in a wide span or recognition tasks, there exists growing evidence suggesting that these models are poorly calibrated, leading to overconfident predictions that may assign high confidence to incorrect predictions \citep{gal2016dropout,guo2017calibration}. 
This represents a major problem, as inaccurate uncertainty estimates can carry serious implications in safety-critical applications such as medical diagnosis, whose outcomes are used in subsequent tasks of critical importance. 
The underlying cause of miscalibration in deep models is hypothesized to stem from their high capacity, which makes them prone to overfitting on the negative log-likelihood loss, commonly used during training \citep{guo2017calibration}. \rev{Indeed, modern classification networks trained under the fully supervised learning paradigm resort to binary one-hot encoded vectors as supervisory signals of training data points. Therefore, all the probability mass is assigned to a single class, resulting in minimum-entropy supervisory signals (i.e., entropy equal to zero). As the network is trained to follow this distribution, we are implicitly forcing it to be overconfident (i.e., to achieve a minimum entropy), thereby penalizing uncertainty in the predictions. }

In light of the significance of this issue, there has been a surge in popularity for quantifying the predictive uncertainty in modern DNNs. A simple approach involves a post-processing step that modifies the softmax probability predictions of an already trained network \citep{guo2017calibration,tomani2021post,zhang2020mix,ding2021local}. \rev{These methods, however, see their performance degrade under distributional drifts \citep{ovadia2019can}. More principled alternatives incorporate a term that maximizes the Shannon entropy of the model predictions during training, penalizing confident output distributions. This regularization term is either implicitly derived from the original loss \citep{mukhoti2020calibrating,muller2019does} or explicitly integrated as additional learning objectives \citep{pereyra2017regularizing,liu2022devil,liu2023class}. }

\rev{Due to the importance of correctly modeling the uncertainty estimates in deep segmentation models, just a few works have recently studied the impact of existing approaches in this problem \citep{jena2019bayesian,larrazabal2021orthogonal,ding2021local,murugesan2022calibrating}. Nevertheless, these approaches are directly borrowed from the classification literature, which presents important limitations in the segmentation scenario. In particular, dense prediction tasks, such as image segmentation, greatly benefit from capturing pixel relationships due to the ambiguity in the boundaries between neighboring organs or regions. Indeed, the nature of structured predictions in segmentation involves pixel-wise classification based on spatial dependencies, which limits the effectiveness of these strategies to yield performances similar to those observed in classification tasks \citep{mukhoti2020calibrating,muller2019does,liu2022devil}. This potentially suboptimal performance can be attributed to the uniform (or near-to-uniform) distribution 
enforced on the softmax/logits distributions, which disregards the spatial context information. While modeling these pixel-wise relationships, for example, by modeling the class distribution around a given pixel, is extremely important, virtually none of existing methods explicitly considers these relationships. }

To address this important issue, Spatially Varying Label Smoothing (SVLS) \citep{islam2021spatially} introduces a soft labeling approach that captures the structural uncertainty required in semantic segmentation. 
In practice, smoothing the hard-label assignment is achieved through a Gaussian kernel applied across the one-hot encoded ground truth, which results in soft class probabilities based on neighboring pixels. Nevertheless, while the reasoning behind this smoothing strategy relies on the intuition of giving an equal contribution to the central label and all surrounding labels combined, its impact on the training, from an optimization standpoint, has not been studied. 

We can summarize our \textbf{contributions} as follows:

\begin{itemize}
    \item In this work, we provide a constrained-optimization perspective of Spatially Varying Label Smoothing (SVLS) \citep{islam2021spatially}, demonstrating that it could be viewed as a standard cross-entropy loss coupled with an implicit constraint that enforces the softmax predictions to match a soft class proportion of surrounding pixels. 
    Our formulation shows that SVLS lacks a 
    mechanism to control explicitly the importance of the constraint, which may hinder the optimization process as it becomes challenging to balance the constraint with the primary objective effectively. 

    \item Following these observations, we propose a simple and flexible solution based on equality constraints on the logit distributions. The proposed constraint is enforced with a simple linear penalty, which incorporates an explicit mechanism to control the weight of the penalty. Our approach not only offers a more efficient strategy to model the logit distributions but implicitly decreases the logit values, which results in less overconfident predictions. 

    \item We conduct comprehensive experiments and ablation studies over multiple medical image segmentation benchmarks, including diverse targets and modalities, and show the superiority of our method compared to state-of-the-art calibration losses. Furthermore, several ablation studies empirically validate the design choices of our approach, as well as demonstrate its model agnostic nature. 
    
\end{itemize}

\rev{This journal version provides a substantial extension of the preliminary work presented in \citep{murugesan2023trust}. More concretely, we first provide a thorough literature review on calibration models, with an extensive overview of their use in medical image segmentation. Second, we perform a  comprehensive empirical validation, including \textit{i)} multiple additional public benchmarks covering diverse modalities and targets, 
\textit{ii)} several ablation studies that motivate our choices, \textit{iii)} showing the agnostic nature of NACL regarding the segmentation backbone, and \textit{iv)} additional results that help us to understand the underlying benefits of the proposed approach.}

\section{Related work}
\label{sec:related}

\noindent \textbf{Post-processing approaches.} A straightforward and effective approach to mitigate the miscalibration issue involves implementing a post-processing step that transforms the probability predictions of a deep network \citep{guo2017calibration, zhang2020mix, tomani2021post}. In this scenario, a validation set, drawn from the generative distribution of the training data $\pi (X, Y)$ is leveraged to rescale the network outputs, resulting in well-calibrated in-domain predictions. Temperature scaling (TS) \citep{guo2017calibration}, a simple generalization of Platt scaling \citep{platt1999probabilistic} to the multi-class setting, 
uses a single value overall logit (i.e., pre-softmax) predictions to control the shape of the class predicted distributions. 
\citep{tomani2021post} proposes to transform the validation set before transforming the softmax distributions, whereas \citep{zhang2020mix} combines isotonic regression (IR) after performing temperature scaling.  
Despite its efficiency, most approaches within this family present important limitations, including \textit{i)} a dataset-dependency on the value of the transformation parameters 
and \textit{ii)} a significant degradation observed on out-of-domain samples \citep{ovadia2019can}.

\rev{\noindent \textbf{Penalizing low-entropy predictions.} To alleviate the issue of overconfident predictions inherent in minimizing a negative log-likelihood loss, a natural strategy is to encourage high-entropy, i.e., uncertain, predictions. A straightforward solution to achieve this is to include into the learning objective a term to penalize confident output distributions by explicitly maximizing the entropy \citep{pereyra2017regularizing}. More recently, several works \citep{muller2019does,mukhoti2020calibrating} have shed light into the implicit calibration properties of popular losses (label smoothing and focal loss) that modify the one-hot encoding labels used for training. More concretely, label smoothing \citep{szegedy2016rethinking} has been shown to implicitly calibrate the trained models, as it prevents the network from assigning the full probability mass to a single class, while encouraging the differences between the logits of the target class and the other categories to be a constant dependent on $\alpha$\footnote{In label smoothing, $\alpha$ controls the mass that is uniformly distributed across the different classes: $y^{LS}_k=y_k (1-\alpha)+\alpha/K$.} \citep{muller2019does}. In addition, \citep{mukhoti2020calibrating} demonstrated that focal loss \citep{lin2017focal} implicitly minimizes a Kullback-Leibler (KL) divergence between the uniform distribution and the softmax network predictions, thereby increasing the entropy of the predictions. Thus, we can see both label smoothing and focal loss as classification losses that implicitly regularize the network output probabilities, encouraging their distribution to be close to the uniform distribution. More recently, \citep{liu2022devil} presented a unified view of state-of-the-art calibration approaches \citep{pereyra2017regularizing,szegedy2016rethinking,lin2017focal} 
showing that these strategies can be viewed as approximations of a linear penalty enforcing equality constraints on logit distances, which are encouraged to be zero across all the logits. This view exposes important limitations of the ensuing gradients, which constantly push towards a non-informative solution, compromising an optimal trade-off between discriminative and calibration performance. To circumvent this limitation, authors proposed a simple and flexible generalization of label smoothing (MbLS) based on inequality constraints, which imposes a controllable margin on logit distances. }

\rev{\paragraph{\textbf{Calibration in medical image segmentation}} Despite recent efforts to model the predictive uncertainty, or to leverage this uncertainty to improve the discriminative performance of segmentation models \citep{wang2019aleatoric}, little attention has been devoted to improving both the calibration and segmentation performance of deep models in the medical domain. 
\citep{jena2019bayesian} presented a Bayesian decision theoretic framework based on deep models for image segmentation. This framework produced analytical estimates of uncertainty, allowing to define a principled measure of uncertainty associated with label probabilities, which led to an improvement on both segmentation and calibration performances. Nevertheless, there exists recent evidence \citep{fort2019deep} that indicates that Bayesian neural networks tend to find solutions around a single minimum of the loss landscape, resulting in a lack of diversity. In contrast, ensembling multiple deep neural networks usually yields more diverse predictions, consequently leading to improved uncertainty estimates which outperform other methods \citep{jungo2020analyzing,mehrtash2020confidence}. In the context of medical image segmentation, several strategies have been adopted to promote model diversity within the ensemble, such as imposing orthogonality constraints during training \citep{larrazabal2021orthogonal} or training a single model in a multi-task manner on several different datasets \citep{karimi2022improving}. A main drawback of these approaches, however, lies in their increased complexity cost, as they entail the training of either multiple models or a single model on multiple datasets.}

\rev{\cite{ding2021local} present a lighter alternative that extends the simple temperature scaling approach by integrating a shallow neural network to predict the voxel-wise temperature values, which are used in a post-processing step. While this method 
outperforms the naive TS, it inherits the limitations of temperature scaling and related post-processing approaches. More recently, \citep{murugesan2022calibrating} performed a comprehensive evaluation of existing calibration approaches in the task of medical image segmentation. The reported results suggested that methods integrating explicit penalties, and in particular MbLS \citep{liu2022devil}, largely outperformed other existing techniques in both discrimination and calibration metrics. All these methods, however, are predominantly adopted from the classification literature, which ignores the underlying properties of dense prediction problems, such as semantic image segmentation. In these cases, the spatial relations between a given pixel and its neighbors play a crucial role in the predictions, and the surrounding class distributions in the pixel-wise annotations should be considered for modeling the uncertainty. 
Indeed, and as to the best of our knowledge, the work in \citep{islam2021spatially} is the only method that considers the pixel vicinity of the labeled mask to improve the calibration performance of deep segmentation models. More concretely, authors apply a Gaussian kernel across the one-hot encoded labels to obtain soft class probabilities, integrating spatial-awareness into the standard label smoothing process. }

\section{Methodology}

\subsection{Preliminaries}

\noindent \textbf{Notation}. 
Let us denote the training dataset as $\mathcal{D}=\{(\vx^{(n)}, \vy^{(n)})\}_{n=1}^N$, \rev{where the set of $N$ pairs are \textit{i.i.d.} realizations of the random variables $X,Y$ which follow a ground truth joint distribution $\pi (X,Y)=\pi(X|Y)\pi(X).$ 
In this setting, $\vx^{(n)} \in \sR^{\Omega_n}$ represents the $n^{th}$ image, $\Omega_n$ the spatial image domain, and $\vy^{(n)} \in \sR^K$ its corresponding ground-truth label with $K$ classes, 
provided as a one-hot encoding vector. For simplicity and clarity in the formulation, we will omit in what follows the superscript to indicate the sample used, and $\vx$ will denote any image in the training set. Now, given an input image $\vx$, a neural network parameterized by $\theta$ generates the set of logit predictions $f_{\theta}(\vx)=\rvl \in \sR^{\Omega_n \times K}$. Last, we use the softmax function, denoted as $\phi (\cdot)$ to obtain the predicted model probabilities $\hat{\vp}=\phi(f_{\theta}(\vx)) \in \sR^{\Omega_n \times K}$.}

\rev{\noindent \textbf{What is calibration?} Calibration measures the correspondence between the predicted probabilities assigned by a model and the empirical likelihood of the associated events. A well-calibrated model ensures that its predicted probabilities align with the actual observed frequencies of outcomes. For instance, when the model assigns a probability of 0.7 to an event, it is expected that this event materializes approximately 70\% of the time in the empirical data. In a classification scenario, we can formally define that a model presents \textit{perfect calibration of confidence} if the following conditional probability holds:}

\begin{equation}
\label{eq:condProb}
\rev{\sP(\hat{y} = y| \hat{p} = p)= p, \quad \forall p \in [0,1],}
\end{equation}

\rev{\noindent where $\hat{y} = \arg \max (\hat{\vp})$ is the predicted class of input image $\vx$, and $\hat{p} = \max (\hat{\vp})$ its associated confidence. Equation \ref{eq:condProb} tells us that, to be perfectly calibrated, when the model predicts the probability distribution $\phi(f_{\theta}(X))$ over the set of classes $[K]=\{1,2,...,K\}$, the true probability distribution for these categories should be $\phi(f_{\theta}(X))$. Thus, any difference between the left and right terms is known as calibration error, or \textit{miscalibration}.}

\subsection{A constrained optimization perspective of SVLS}
\label{ssec:SVLS}

Spatially Varying Label Smoothing (SVLS) \citep{islam2021spatially} considers the surrounding class distribution of a given pixel $p$ in the ground truth $\yy$ to estimate the amount of smoothness over the one-hot label of that pixel. In particular, let us consider that we have a 2D patch $\bf{x}$ of size $d_1 \times d_2$ and its corresponding ground truth $\bf{y}$\footnote{For the sake of simplicity, we consider a patch as an image $\xx$ (or mask $\yy$), whose spatial domain $\Omega$ is equal to the patch size, i.e., $d_1 \times d_2$.}. Furthermore, the predicted softmax in a given pixel is denoted as $\rvs=[s_0,s_1,...,s_{k-1}]$. 
Let us now transform the surrounding patch of the segmentation mask around a given pixel 
into a unidimensional vector $\bf{y} \in \mathbb{R}^d$, where $d=d_1 \times d_2$. SVLS employs a discrete Gaussian kernel $\bf{w}$ to obtain soft class probabilities from one-hot labels, which can also be reshaped into $\bf{w} \in \mathbb{R}^d$. 
Thus, for a given pixel $j$, and a class $k$, SVLS \cite{islam2021spatially} can be defined as:

\begin{align}
\label{eq:svls}
\tilde{y}^k_j = \frac{1}{| \sum_i^d w_i |} \sum_{i=1}^d y^k_i w_i .
\end{align}

We can replace the smoothed labels $\tilde{y}^k_p$ in the standard cross-entropy (CE) loss, resulting in the following learning objective:

\begin{align}
\label{eq:svls-ce}
\mathcal{L} 
= - \sum_k \left(\frac{1}{|  \sum_i^d w_i |}  \sum_{i=1}^d y^k_i w_i \right) \log \hat{p}^k_j ,
\end{align}

where $s^k_p$ is the softmax probability for the class $k$ at pixel $p$ (the pixel in the center of the patch). Now, we can decompose this loss into:

\begin{align}
\label{eq:svls-ce2}
\mathcal{L}  =  & - \frac{1}{| \sum_i^d w_i |} \sum_k   y^k_{j} \log \hat{p}^k_j  \\
&- \frac{1}{|  \sum_i^d w_i |} \sum_k  
\left(\sum_{\substack{i=1\\i\neq j}}^d y^k_i w_i\right) \log \hat{p}^k_j ,
\end{align}
with $p$ denoting the index of the pixel in the center of the patch. 
Note that the term in the left is the 
cross-entropy between the posterior softmax probability and the hard label 
assignment for pixel $p$. Furthermore, let us denote $\tau_k = \sum_{\substack{i=1\\i\neq j}}^d y^k_i w_i$ as the soft proportion of the class $k$ inside the patch/mask $\bf{y}$, weighted by the filter values $\mathbf{w}$. By replacing $\tau_k$ into the Eq. \ref{eq:svls-ce2}, and removing $|\sum_i^d w_i |$ as it multiplies both terms, 
the loss becomes:

\begin{align}
    \label{eq:tau}
    \mathcal{L}  = \underbrace{-\sum_k   y^k_{j} \log \hat{p}^k_j}_{CE} \underbrace{- \sum_k \tau_k \log \hat{p}^k_j}_{\textrm{Constraint on $\bm{\tau}$}}.
\end{align}

As $\bm{\tau}$ is constant, the second term in Eq. \ref{eq:tau} can be replaced by a Kullback-Leibler (KL) divergence, leading to the following learning objective:

\begin{align}
    \label{eq:globalLoss}
    \mathcal{L}  \ceq \mathcal{L}_{CE} + 
    \mathcal{D}_{KL}(\bm{\tau}||\hat{\vp}) ,
\end{align}
where $\ceq$ stands for equality up to additive and/or non-negative multiplicative constant. Thus, optimizing the loss 
in SVLS results in minimizing the 
cross-entropy between the hard label and the softmax probability distribution on the pixel $j$, 
while imposing the equality constraint $\bm{\tau}=\hat{\vp}$, where $\bm{\tau}$ depends on the class distribution of surrounding pixels. 
Indeed, this term 
implicitly enforces the softmax predictions to match the soft-class proportions computed around pixel $j$.

\subsection{Proposed constrained calibration approach}
\label{ssec:ours}

Our previous analysis exposes two important limitations of SVLS: \textit{1)} the importance of the implicit constraint cannot be controlled explicitly, and \textit{2)} the prior $\bm{\tau}$ is derived from the $\sigma$ value in the Gaussian filter, making it difficult to model properly. 
To alleviate this issue, we propose a simple solution, which consists in minimizing the standard cross-entropy between the softmax predictions and the one-hot encoded masks coupled with an explicit and controllable constraint on the logits $\rvl$. In particular, we propose to minimize the following constrained objective:

\begin{align}
\label{eq:proposed-constrained}
\min_{\theta} \quad \mathcal{L}_{CE} \quad \textrm{s.t.} \quad  \bm{\tau} = \rvl, 
\end{align}

\noindent where $\bm{\tau}$ now represents a desirable prior, and $\bm{\tau} = \rvl $ is a hard constraint. Note that the reasoning behind working directly on the logit space is two-fold. First, observations in \cite{liu2022devil} suggest that directly imposing the constraints on the logits 
results in better performance than in the softmax predictions. And second, by imposing a bounded constraint on the logits values\footnote{Note that the proportion priors are generally normalized.}, their magnitudes are further decreased, which has a favorable effect on model calibration \cite{muller2019does}. We stress that despite both \cite{liu2022devil} and our method enforce constraints on the predicted logits, \cite{liu2022devil} is fundamentally different. 
In particular, \cite{liu2022devil} imposes an \textit{inequality} constraint on the logit distances so that it encourages uniform-alike distributions up to a given margin, disregarding the importance of each class in a given patch. 
This can be important in the context of image segmentation, where 
the uncertainty of a given pixel may be strongly correlated with the labels assigned to its neighbors. In contrast, our solution enforces \textit{equality} constraints on an adaptive prior, encouraging distributions close to class proportions in a given patch.     

Even though the constrained optimization problem presented in Eq. \ref{eq:proposed-constrained} could be solved by a standard Lagrangian-multiplier algorithm, we replace the hard constraint by a soft penalty of the form $\mathcal{P}(|\bm \tau - \rvl|)$, transforming our constrained problem into an unconstrained one, which is easier to solve. In particular, the soft penalty $\mathcal{P}$ should be a continuous and differentiable function that reaches its minimum when it verifies $\mathcal{P}(|\bm \tau - \rvl|) \geq \mathcal{P}(\bm 0),  \, \forall \, \bm l \in \mathbb{R}^{K}$, i.e., when the constraint is satisfied. Following this, when the constraint $|\bm \tau - \rvl|$ deviates from $\bm 0$ the value of the penalty term increases. Thus, we can approximate the problem in Eq. \ref{eq:proposed-constrained} as the following simpler unconstrained problem:


\begin{align}
\label{eq:proposed-unconstrained}
\min_{\theta} \quad \mathcal{L}_{CE} + \lambda \sum_k |\tau_k - l_k| ,
\end{align}
where the hyperparameter $\lambda$ controls 
the importance of the penalty.

\section{Experiments}
\subsection{\rev{Experimental Setting}}

\subsubsection{\rev{Datasets}}
\rev{To empirically validate our model, we resort to six public multi-class segmentation benchmarks, whose details are provided below.} 

\noindent{\textbf{{\rev{Automated Cardiac Diagnosis Challenge (ACDC)}}} \citep{bernard2018deep}.} 
\rev{This dataset comprises short-axis cardiac cine-MRI scans from 100 patients, in both diastolic and systolic phases with their respective segmentation annotations. The task of this challenge is to understand the cardiac function through segmenting key regions, including the left ventricle (LV), the right ventricle (RV), and the myocardium (Myo). Following standard practices, we randomly split the dataset into 70 patients for training, 10 for validation, and the remaining 20 for testing. From each of these volumes, we extract 2D slices, which are resized to 224$\times$224.}

\noindent{\textbf{\rev{Brain Tumor Segmentation (BRATS) 2019 Challenge}} \citep{Menze2015TheBRATSJ, Bakas2017AdvancingFeaturesJ, Bakas2018IdentifyingChallengeJ}.}
\rev{The goal of this challenge is to identify glioma tumors in multi-channel MRI scans (FLAIR, T1, T1-contrast, and T2). The dataset consists of $335$ volumes with their corresponding segmentation masks, which include tumor core (TC), enhancing tumor (ET), and whole tumor (WT). Following prior works, we randomly split the volumes into subsets of 235, 35, and 65 scans for training, validation, and testing, respectively. We also resample the volumes, extract the 2D slices and discard the empty slices.}

\noindent{\textbf{\rev{Fast and Low GPU memory Abdominal oRgan sEgmentation (FLARE) Challenge}} \citep{ma2021abdomenct}.} 
\rev{
This dataset contains $360$ abdominal CT scans obtained from diverse medical centers with pixel-wise masks of several organs, including liver, kidneys, spleen, and pancreas. Following standard protocols, we randomly split the scans into 240 for training, 40 for validation, and 80 for testing. Furthermore, CT scans with different resolutions are resampled to the same space and cropped to 192$\times$192$\times$30, from which 2D slices are obtained.}

\noindent{\textbf{\rev{PROSTATE}}} \citep{antonelli2022medical} \rev{The dataset was acquired at Radboud University Medical Center and was released as a part of Medical Segmentation Decathlon (MSD) challenge. The dataset consists of 32 MRI volumes with target regions of prostate peripheral zone (PZ) and the transition zone (TZ). The dataset is challenging because of segmenting two adjoined regions large inter-subject variability. We split the dataset to $22$ patients for training, $3$ for validation and $7$ for testing.}

\noindent{\textbf{\rev{KIdney Tumor Segmentation (KiTS) challenge}} \citep{heller2019kits19}.}  
\rev{
This dataset consists of 210 CT scans with their respective segmentation masks, including the kidney and tumor classes. Following \citep{islam2021spatially}, we resampled cases with varying resolutions and image sizes to a common resolution of 3.22 $\times$ 1.62 $\times$ 1.62 mm and center crop to image size 80 $\times$ 160 $\times$ 160. The dataset is randomly split into 150 cases for training, 25 for validation, and 40 for testing.}

\noindent{\textbf{\rev{MRBrainS18}} \citep{mendrik2015mrbrains}.} 
\rev{The purpose of this challenge is to segment the brain MRI scans into  Gray Matter (GM), White Matter (WM), and Cerebralspinal fluid (CSF). 
The dataset contains paired T1, T2, and T1-IR sequences of 3D volumes (240$\times$240$\times$48) of 7 subjects and their associated pixel-wise masks. For the experiments, we consider 5 subjects for training and 2 for testing.} 

\rev{Note that in all the datasets, images are normalized to be within the range [0-1]. Furthermore, for the datasets containing multiple image modalities (i.e., MRBrainS and BraTS), all available modalities are concatenated in a single tensor, which is fed to the input of the neural network. In addition, there exists one dataset for which the low amount of available images impeded us to generate a proper training, validation, and testing split (MRBrainS). In this case, we performed leave-one-out-cross-validation in our experiments, whereas the other datasets followed standard training, validation, and testing procedures, using a single split in the experiments.}

\subsubsection{Evaluation Metrics}
\rev{We assess the discriminative performance of the model using standard segmentation metrics in the medical imaging community, including the overlap-based metric DICE (DSC) coefficient, and spatial distance metric Hausdorff distance (HD). For understanding the calibration performance, we resort to Expected Calibration Error (ECE) and Classwise Expected Calibration Error (CECE) \citep{naeini2015obtaining}. ECE concentrates only on maximum confidence score of the prediction, while CECE considers the confidence distribution of all the classes, including the winner class \citep{mukhoti2020calibrating}. Importantly, we obtain the calibration metrics only for the foreground regions following the recent literature   \citep{islam2021spatially,murugesan2022calibrating}. The notion behind this is because the class distribution is skewed towards background, particularly in most cases of medical image segmentation. Hence, excluding background allows us to better compare the performance of different methods. We further understand the calibration performance through reliability plots \citep{niculescu2005predicting}, wherein accuracy is expected to be directly correlated to class probability. In both the cases, we set the number of bins to $15$.}

\rev{To compute ECE and CECE for $N$ samples with $K$ classes, we group predictions into $M$ equispaced bins.  Let $B_{i}$ denote the set of samples with maximum confidences belonging to the $i^{th}$ bin, and $B_{ij}$ denotes the set of samples from the $j^{th}$ class in the $i^{th}$ bin. The accuracy $A_{i}$ of  $i$-th bin is computed as $A_{i}=\frac{1}{|B_{i}|}\sum_{j \in B_{i}}1(\hat{y_{j}} =y_{j})$, where 1 is the indicator function. Similarly, for class-wise, the accuracy is given by $A_{i,j}=\frac{1}{|B_{i,j}|}\sum_{k \in B_{i,j}}1(j =y_{k})$. The confidence $C_{i}$ of the $i^{th}$ bin and $C_{i,j}$ of $i^{th}$ bin, $j^{th}$ class is given by $C_{i}=\frac{1}{|B_{i}|}\sum_{j \in B_{i}}\hat{p}_{j}$ and $C_{i,j}=\frac{1}{|B_{ij}|}\sum_{k \in B_{i,j}}\hat{p}_{kj}$ respectively. Hence, ECE and CECE is given by: 


\begin{align}
\label{eq:ece}
ECE = \sum_{i=1}^{M}\frac{|B_{i}|}{N}|A_{i} - C_{i}|
\end{align}

\begin{align}
\label{eq:cece}
CECE = \sum_{i=1}^{M}\sum_{j=1}^{K}\frac{|B_{i,j}|}{N}|A_{i,j} - C_{i,j}|
\end{align}
}

\subsubsection{Implementation Details}
\rev{To empirically evaluate the proposed model, we conduct experiments comparing a state-of-the-art segmentation network on a multi-class scenario trained with different learning objectives. In particular, we first employ standard loss functions employed in medical image segmentation, which include the popular Cross-entropy (CE) combined with DSC loss. Furthermore, we also include training objectives that have been proposed to calibrate deep neural networks for both classification and segmentation problems, which represent nowadays the state-of-the-art for this task. This includes Focal loss (FL) \citep{lin2017focal}, Label Smoothing (LS) \citep{szegedy2016rethinking}, ECP  \citep{pereyra2017regularizing}, SVLS \citep{islam2021spatially}, and MbLS \citep{liu2022devil}. Following the literature, we have chosen the following hyper-parameters for the different approaches: FL ($\gamma$=3.0), ECP ($\alpha$=0.1), LS ($\lambda$=0.1), SVLS ($\sigma$=2.0) and MbLS ($m$=5). Note that in the main experiments, these hyperparameters remain fixed across the different datasets for all the models, to better highlight the generability of each approach. For the experiments, we fixed the batch size to 16, epochs to 100, and optimizer to ADAM. The learning rate of 1e-3 and 1e-4 are used for the first 50 epochs, and the next 50 epochs, respectively. The models are trained on 2D slices, while the evaluation is done over 3D volumes. The best model is selected based on the mean DSC score on the validation dataset. }

\noindent \textbf{Backbones.} The experiments are predominantly conducted on the standard UNet \citep{unet} architecture. \rev{Nevertheless, to demonstrate the model-agnostic nature of our approach we also evaluate the effect of our method on other common architectures in medical image segmentation, including convolutional neural networks (AttUNet \citep{attunet}, UNet++ \citep{unetpp} and nnUNet \citep{isensee2021nnu}) and Vision Transformer based architectures (TransUNet \citep{transunet}).}

\begin{table*}[h!]
\scriptsize
\centering
\caption{Discriminative performance obtained by the different evaluated models across six popular medical image segmentation benchmarks. Best method is highlighted in bold, whereas second best approach is underlined.}
\label{tab:main-disc}
\begin{tabular}{c|c|cccccccccccccc}
\toprule
Dataset &  Region & \multicolumn{2}{c}{CE+DSC} & \multicolumn{2}{c}{FL} & \multicolumn{2}{c}{ECP} & \multicolumn{2}{c}{LS} & \multicolumn{2}{c}{SVLS} & \multicolumn{2}{c}{MbLS} & \multicolumn{2}{c}{\ours}\\
\midrule
& & DSC $\uparrow$ & HD $\downarrow$ & DSC $\uparrow$ & HD $\downarrow$  & DSC $\uparrow$ & HD $\downarrow$  & DSC $\uparrow$ & HD $\downarrow$  & DSC $\uparrow$ & HD $\downarrow$  & DSC $\uparrow$ & HD $\downarrow$  & DSC $\uparrow$ & HD $\downarrow$  \\
\midrule
\multirow{4}{*}{ACDC} & RV 
 & 0.799 & 3.10 & 0.580 & 9.37 & 0.751 & 4.93 & 0.796 & 3.34 & 0.791 & \und{2.89} & \und{0.812} & \textbf{2.59} & \textbf{0.837} & 3.02 \\ 
 & MYO & 0.795 & \und{2.57} & 0.557 & 5.55 & 0.757 & 3.54 & 0.772 & 3.07 & \und{0.798} & 2.66 & 0.795 & 2.86 & \textbf{0.820} & \textbf{2.04} \\ 
 & LV & \und{0.889} & 3.75 & 0.724 & 6.97 & 0.839 & 4.85 & 0.858 & 3.49 & 0.882 & \und{2.89} & 0.875 & 3.53 & \textbf{0.905} & \textbf{2.59} \\ 
 \rowcolor{gray!20} & Mean & \und{0.828} & 3.14 & 0.620 & 7.30 & 0.782 & 4.44 & 0.809 & 3.30 & 0.824 & \und{2.81} & 0.827 & 2.99 & \textbf{0.854} & \textbf{2.55}\\  
 \midrule
 \multirow{5}{*}{FLARE} & Liver & 0.950 & \und{6.09} & 0.952 & 7.54 & \und{0.953} & 7.41 & 0.952 & 8.50 & 0.951 & 7.72 & 0.941 & 7.18 & \textbf{0.954} & \textbf{6.04} \\
 & Kidney & 0.945 & 2.07 & 0.947 & 2.16 & \und{0.950} & 2.05 & 0.947 & \textbf{1.76} & 0.947 & 1.84 & 0.937 & 2.49 & \textbf{0.952} & \und{1.84} \\
 & Spleen & 0.892 & 9.49 & 0.887 & 9.09 & 0.887 & \textbf{3.98} & \textbf{0.905} & 4.62 & 0.879 & 6.40 & 0.868 & 4.73 & \und{0.900} & 4.26\\
 & Pancreas & 0.636 & 7.95 & 0.626 & 7.80 & 0.649 & 7.77 & 0.637 & 6.45 & \und{0.650} & 6.91 & 0.596 & 8.61 & \textbf{0.664} & 7.37\\
 \rowcolor{gray!20} & Mean & 0.855 & 6.40 & 0.853 & 6.65 & \und{0.860} & 5.30 & 0.860 & 5.33 & 0.857 & 5.72 & 0.836 & 5.75 & \textbf{0.867} & \textbf{4.88}\\
 \midrule
\multirow{4}{*}{BraTS} & TC & 0.731 & 5.73 & 0.799 & 7.80 & 0.749 & 7.53 & 0.773 & 5.16 & 0.744 & 7.56 & \und{0.803} & 4.88 & \textbf{0.804} & \textbf{3.98}\\ 
 & ET & 0.766 & 8.27 & \und{0.854} & 10.02 & 0.790 & 11.31 & 0.807 & 10.23 & 0.783 & 9.22 & 0.821 & 10.85 & \textbf{0.854} & \textbf{6.58}\\
& WT & 0.872 & 6.88 & 0.889 & 9.19 & 0.884 & 7.28 & 0.879 & 7.94 & 0.877 & 8.55 & \und{0.889} & 8.09 & \textbf{0.893} & \textbf{6.78}\\ 
\rowcolor{gray!20} & Mean & 0.789 & 6.96 & \und{0.848} & 9.00 & 0.808 & 8.71 & 0.820 & 7.78 & 0.801 & 8.44 & 0.838 & 7.94 & \textbf{0.850} & \textbf{5.78}\\  
 \midrule
 \multirow{3}{*}{\rev{PROSTATE}} & \rev{CG} & \rev{0.329} & \rev{16.00} & \rev{0.223} & \rev{23.45} & \rev{0.344} & \rev{19.97} & \rev{0.292} & \rev{13.51} & \rev{0.341} & \rev{15.24} & \rev{\textbf{0.427}} & \rev{\textbf{10.93}} & \rev{\und{0.418}} & \rev{\underline{12.73}} \\ 
 & \rev{PZ} & \rev{0.752} & \rev{7.13} & \rev{0.677} & \rev{12.57} & \rev{0.736} & \rev{6.19} & \rev{0.756} & \rev{\underline{5.12}} & \rev{0.737} & \rev{9.28} & \rev{\und{0.774}} & \rev{5.65} & \rev{\textbf{0.796}} & \rev{\textbf{4.02}} \\
\rowcolor{gray!20} & \rev{Mean} & \rev{0.540} & \rev{11.56} & \rev{0.450} & \rev{18.01} & \rev{0.540} & \rev{13.08} & \rev{0.524} & \rev{9.31} & \rev{0.539} & \rev{12.26} & \rev{\und{0.601}} & \rev{\textbf{8.29}} & \rev{\textbf{0.607}} & \rev{\underline{8.37}} \\  
\midrule
\multirow{3}{*}{\rev{KiTS}} & \rev{Kidney} & \rev{\textbf{0.786}} & \rev{9.11} & \rev{\und{0.784}} & \rev{\textbf{8.74}} & \rev{0.735} & \rev{10.27} & \rev{0.759} & \rev{\underline{9.06}} & \rev{0.770} & \rev{9.86} & \rev{0.749} & \rev{10.56} & \rev{0.780} & \rev{9.08} \\ 
& \rev{Tumor} & \rev{0.447} & \rev{\textbf{13.09}} & \rev{\und{0.470}} & \rev{\und{13.57}} & \rev{0.365} & \rev{15.49} & \rev{0.446} & \rev{16.61} & \rev{0.468} & \rev{15.96} & \rev{0.426} & \rev{16.85} & \rev{\textbf{0.525}} & \rev{15.77} \\
\rowcolor{gray!20} & \rev{Mean} & \rev{0.616} & \rev{\textbf{11.10}} & \rev{\und{0.627}} & \rev{\und{11.15}} & \rev{0.550} & \rev{12.88} & \rev{0.602} & \rev{12.83} & \rev{0.619} & \rev{12.91} & \rev{0.588} & \rev{13.71} & \rev{\textbf{0.652}} & \rev{12.42} \\  
\midrule
\multirow{4}{*}{\rev{MRBrainS}} & \rev{GM} & \rev{\und{0.754}} & \rev{1.73} & \rev{0.672} & \rev{2.81} & \rev{0.747} & \rev{2.23} & \rev{0.707} & \rev{2.12} & \rev{0.725} & \rev{\underline{1.71}} & \rev{0.741} & \rev{2.09} & \rev{\textbf{0.781}} & \rev{\textbf{1.41 }}\\ 
& \rev{WM} & \rev{0.759} & \rev{2.91} & \rev{0.598} & \rev{5.60} & \rev{\und{0.783}} & \rev{\underline{2.73}} & \rev{0.702} & \rev{4.98} & \rev{0.603} & \rev{6.24} & \rev{0.729} & \rev{3.08} & \rev{\textbf{0.791}} & \rev{\textbf{2.64}} \\
& \rev{CSF} & \rev{0.776} & \rev{2.00} & \rev{0.722} & \rev{4.18} & \rev{0.746} & \rev{3.10} & \rev{0.730} & \rev{2.34} & \rev{\und{0.800}} & \rev{\und{1.41}} & \rev{0.769} & \rev{1.71} & \rev{\textbf{0.820}} & \rev{\textbf{1.21}} \\ 
\rowcolor{gray!20} & \rev{Mean} & \rev{\und{0.763}} & \rev{\und{2.22}} & \rev{0.664} & \rev{4.20} & \rev{0.759} & \rev{2.68} & \rev{0.713} & \rev{3.15} & \rev{0.709} & \rev{3.12} & \rev{0.747} & \rev{2.29} & \rev{\textbf{0.797}} & \rev{\textbf{1.75}} \\  
\bottomrule
\end{tabular}
\end{table*}

\begin{table*}[h!]
\scriptsize
\centering
\caption{Calibration performance obtained by the different evaluated models across six popular medical image segmentation benchmarks. Best method is highlighted in bold, whereas second best approach is underlined. In this case, the calibration metrics are averaged across the different target objects.}
\label{tab:main-cal}
\begin{tabular}{c|cccccccccccccc}
\toprule
Dataset & \multicolumn{2}{c}{CE+DSC} & \multicolumn{2}{c}{FL} & \multicolumn{2}{c}{ECP} & \multicolumn{2}{c}{LS} & \multicolumn{2}{c}{SVLS} & \multicolumn{2}{c}{MbLS} & \multicolumn{2}{c}{\ours}\\ 
\midrule
& ECE $\downarrow$ & CECE $\downarrow$ & ECE $\downarrow$ & CECE $\downarrow$ & ECE $\downarrow$ & CECE $\downarrow$  & ECE $\downarrow$ & CECE $\downarrow$  & ECE $\downarrow$ & CECE $\downarrow$  & ECE $\downarrow$ & CECE $\downarrow$ & ECE $\downarrow$ & CECE $\downarrow$  \\
\midrule
ACDC & 0.137 & 0.084 & 0.153 & 0.179 & 0.130 & 0.094 &\und{0.083} & 0.093 & 0.091 & 0.083 & 0.103 & \und{0.081} & \textbf{0.048} & \textbf{0.061} \\ 
FLARE & 0.058 & 0.034 & 0.053 & 0.059 & \und{0.037} & \textbf{0.027} & 0.055 & 0.049 & 0.039 & 0.036 & 0.046 & 0.041 & \textbf{0.033} & \underline{0.031} \\ 
BraTS & 0.178 & 0.122 & \textbf{0.097} & 0.119 & 0.139 & 0.100 & 0.112 & 0.108 & 0.146 & 0.111 & 0.127 & \textbf{0.095} & \underline{0.112} & \underline{0.097} \\ 
\rev{PROSTATE} & \rev{0.430} & \rev{0.304} & \rev{\und{0.271}} & \rev{0.381} & \rev{0.306} & \rev{\und{0.252}} & \rev{0.304} & \rev{0.301} & \rev{0.335} & \rev{0.272} & \rev{0.322} & \rev{\textbf{0.250}} & \rev{\textbf{0.253}} & \rev{0.254} \\ 
\rev{KiTS} & \rev{0.188} & \rev{0.144} & \rev{\und{0.098}} & \rev{\und{0.133}} & \rev{0.155} & \rev{0.151} & \rev{0.122} & \rev{0.141} & \rev{0.163} & \rev{0.144} & \rev{0.155} & \rev{0.147} & \rev{\textbf{0.090}} & \rev{\textbf{0.124}} \\ 
\rev{MRBrainS} & \rev{0.177} & \rev{0.105} & \rev{0.085} & \rev{0.123} & \rev{0.084} & \rev{0.082} & \rev{\und{0.061}} & \rev{0.101} & \rev{0.077} & \rev{\und{0.080}} & \rev{0.107} & \rev{0.093} & \rev{\textbf{0.027}} & \rev{\textbf{0.056}} \\ 
\bottomrule
\end{tabular}
\end{table*}

\begin{figure*}[h!]
     \centering
     \begin{subfigure}[b]{\linewidth}
         \centering
         \includegraphics[width=0.9\linewidth]{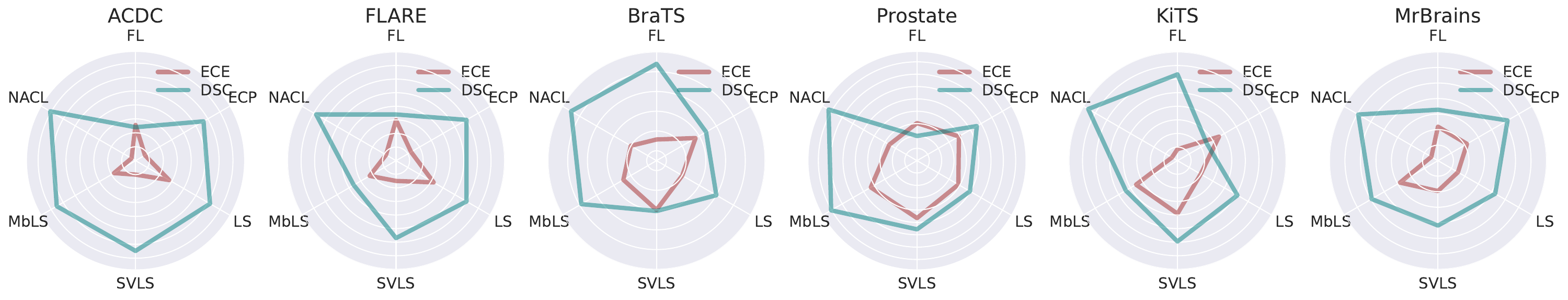}
     \end{subfigure}
    \caption{\rev{\textbf{Compromise between calibration and discriminative performance.} For each dataset, we show the discriminative (DSC) and calibration (ECE) results obtained by each method. 
    We expect a \textit{well-calibrated} model to achieve simultaneously large DSC (\textit{in blue}) and small ECE (\textit{in brown}) values.}}
    \label{fig:radarplot-compromise}
\end{figure*}

\begin{figure}[h!]
     \centering
     \begin{subfigure}[b]{0.49\linewidth}
         \centering
         \includegraphics[width=\linewidth]{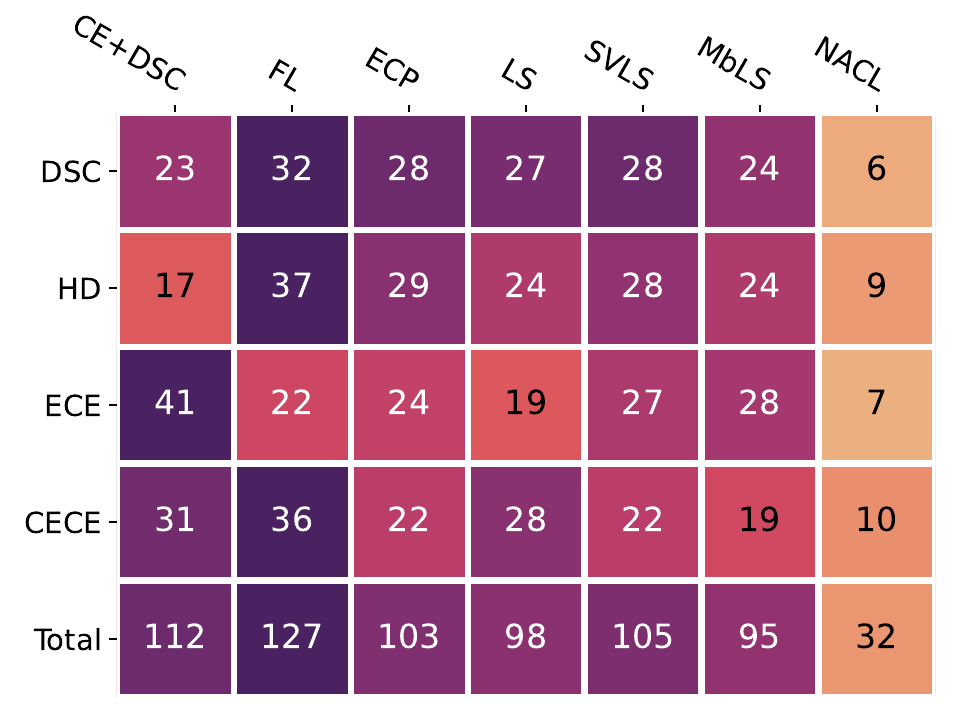}
         \caption{}
         \label{fig:sum-rank}
     \end{subfigure}
     \begin{subfigure}[b]{0.49\linewidth}
         \centering
         \includegraphics[width=\linewidth]{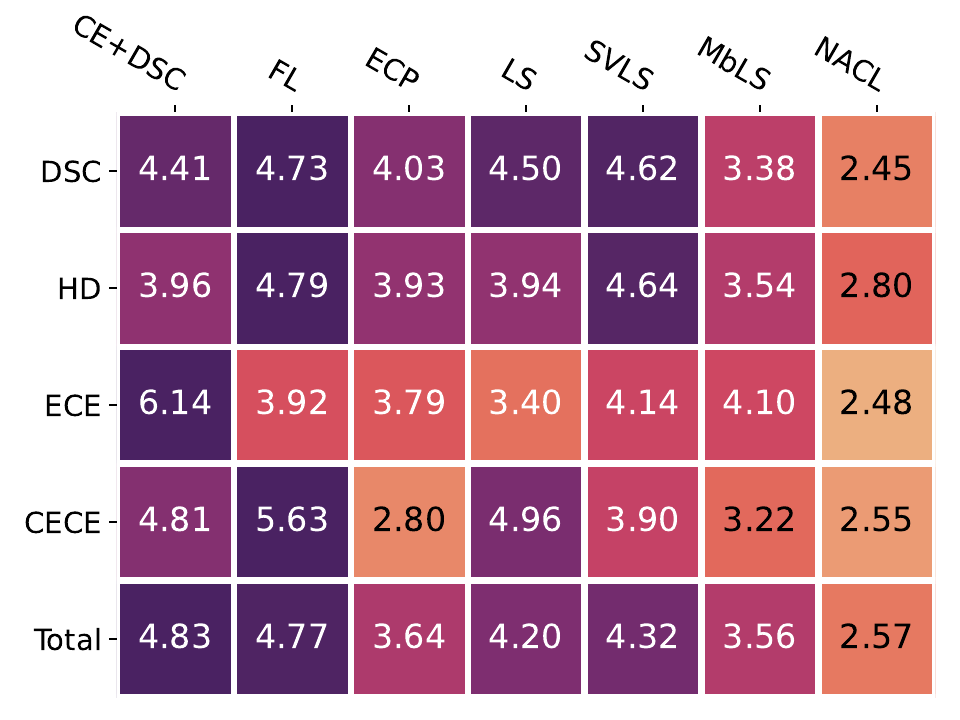}
         \caption{}
         \label{fig:mean-case-rank}
     \end{subfigure}
    \caption{Ranking \textit{global} and \textit{per-metric} of the different methods based on the sum-rank and mean of case-specific approach.}
     \label{fig:rank_figure}
\end{figure}

\subsection{Results}

\subsubsection{Main results}

\rev{We present the quantitative results across a diverse set of segmentation datasets, which include multiple organs, pathologies, as well as several imaging protocols, from a segmentation and a calibration standpoint.}

\rev{\noindent \textbf{Segmentation results.} First, in Table \ref{tab:main-disc}, we compare the discriminative performance of our Neighbor Aware CaLibration method, which we refer to as NACL, to relevant calibration approaches. Notably, we can observe that our approach consistently outperforms existing literature across nearly all the datasets and metrics, yielding improvements which range between 3.4\% and 10\% (DSC), compared to the second and last performing method, respectively. Indeed, if we consider the mean DSC and HD values for each dataset, the proposed approach achieves the best performance in 10 out of the 12 settings, being the second and third best performance method in the remaining 2 scenarios. An important observation is that, whereas our method typically ranks first and second for all targets and metrics, there is no other approach that presents a consistent trend on performance across datasets. For example, Focal loss yields the second best average DSC performance in BraTS, while it ranks last in ACDC or MRBrainS.}    

\rev{\noindent \textbf{Calibration performance.} Similarly to the segmentation scenario, the results in terms of calibration (Table \ref{tab:main-cal}) reveal that our approach consistently yields the best, and second best, uncertainty estimates across datasets and target objects. Furthermore, and as observed in Table \ref{tab:main-disc}, there is no a clear trend on the prior literature, as methods performing competitively in one dataset considerably fail in another, whose discrepancies can also be observed across metrics. For instance, Focal loss yields the best calibrated model, in terms of ECE, for the BraTS dataset, but its ECE value in ACDC is three times higher than the ECE obtained by our approach. This phenomenon is also observed in other approaches, such as ECP (best CECE in FLARE and worst in KiTS) or MbLS (best CECE in BraTS and PROSTATE, but among the worst in MRBrainS). It is important to note that these methods contain different hyperparameters that remained fixed across datasets 
(e.g., $\alpha$ in LS, $\gamma$ in FL, or $\lambda$ and margin $m$ in MbLS). Thus, even though a specific per-dataset fine-tuning of these hyperparameters may lead to a performance increase (both in terms of segmentation and calibration), results in Table \ref{tab:main-disc} and \ref{tab:main-cal} demonstrate empirically that our approach presents a robust alternative to existing methods, as it yields the overall best performance across diverse target objects and datasets.}

\rev{For a more comprehensive understanding of the overall performance across various approaches and datasets, we now introduce two studies that expand upon the quantitative values provided in Table \ref{tab:main-disc} and \ref{tab:main-cal}. First, we resort to radar plots in Figure \ref{fig:radarplot-compromise} to better highlight the trade-off between discriminative and calibration performance achieved by different methods. For a model to be \textit{well-calibrated}, it should present high discriminative performance (\textit{blue line}), while yielding low calibration values (\textit{brown line}). In the case of these radar plots, this implies that a greater distance between the blue and brown lines indicates a more favorable balance between discriminative and calibration performance. Looking at the plots, we can easily observe that the proposed method consistently yields the best trade-offs across datasets, offering high discriminative power without degrading its calibration performance. Other methods, however, must sometimes compromise their discriminative performance to produce calibrated models, or vice-versa. The second study considers the evaluation strategies adopted in several MICCAI Challenges, i.e., sum-rank \citep{mendrik2015mrbrains} and mean-case-rank \citep{MAIER2017250}. As we can observe in the heatmaps provided in Fig. \ref{fig:rank_figure}, our approach yields the best rank across all the metrics in both strategies, clearly outperforming any other method.} Interestingly, some methods such as FL or ECP typically provide well-calibrated predictions, but at the cost of degrading their discriminative performance.

\subsubsection{\rev{On the impact of constraining the logit space}} 

\begin{figure}[h!]
    \centering
    \includegraphics[width=\linewidth]{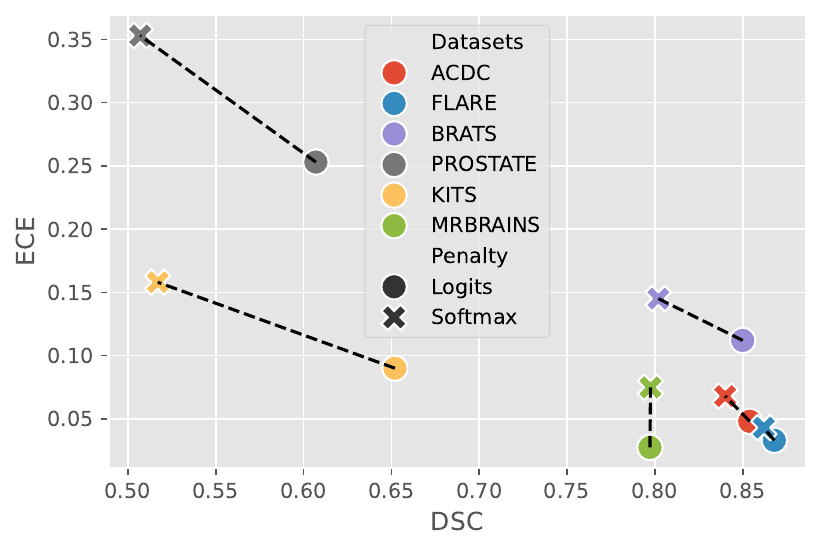}
    \caption{\rev{Impact of applying the penalty over softmax (\textit{cross}) vs logits (\textit{circle}) predictions across the different datasets.}}
    \label{fig:soft-logit-penalty}
\end{figure}

\noindent \textbf{Constraint over logits \textit{vs} softmax.}
Recent evidence \citep{liu2022devil, murugesan2022calibrating} have suggested that imposing constraints on the logits offers a better alternative than its softmax counterpart. To demonstrate that this observation holds in our model, we further present the results of our formulation when the constraint is enforced on the softmax distributions, i.e., replacing $\bf{l}$ by \rev{$\bf{\hat{p}}$ in Equation \ref{eq:proposed-unconstrained}. From these results, reported in Figure \ref{fig:soft-logit-penalty}, it is evident that working on the logit space substantially increases both the segmentation and calibration performance across the datasets. This could be attributed to the range of logits being larger than softmax, allowing for a better control.}

\begin{figure*}[h!]
     \begin{subfigure}[b]{0.19\linewidth}
         \centering
         \includegraphics[width=\linewidth]{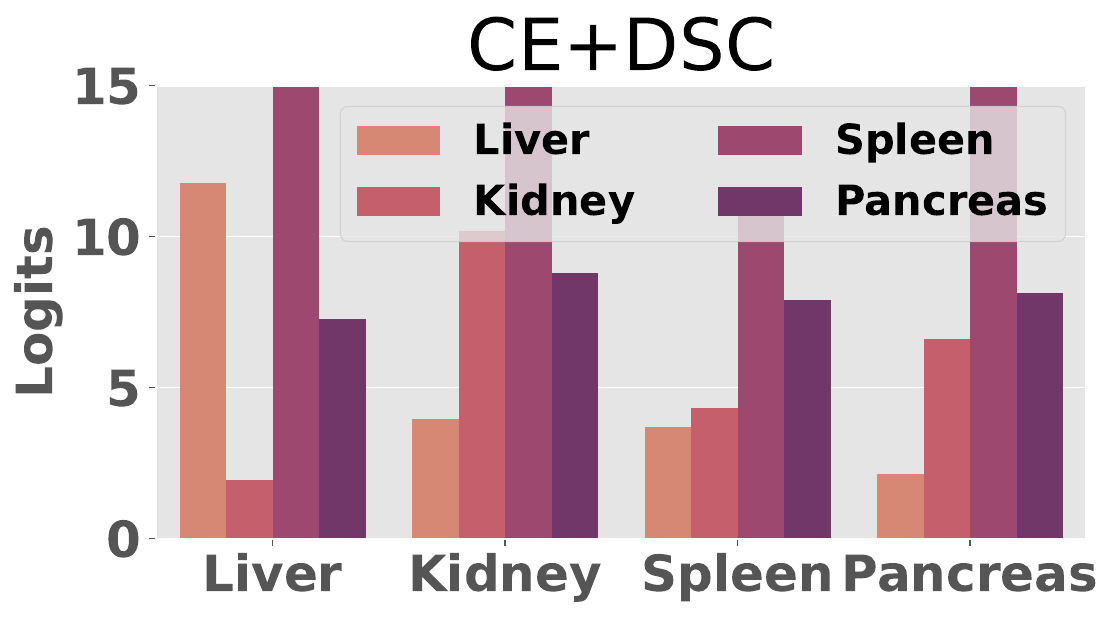}
     \end{subfigure}
     \begin{subfigure}[b]{0.19\linewidth}
         \centering
         \includegraphics[width=\linewidth]{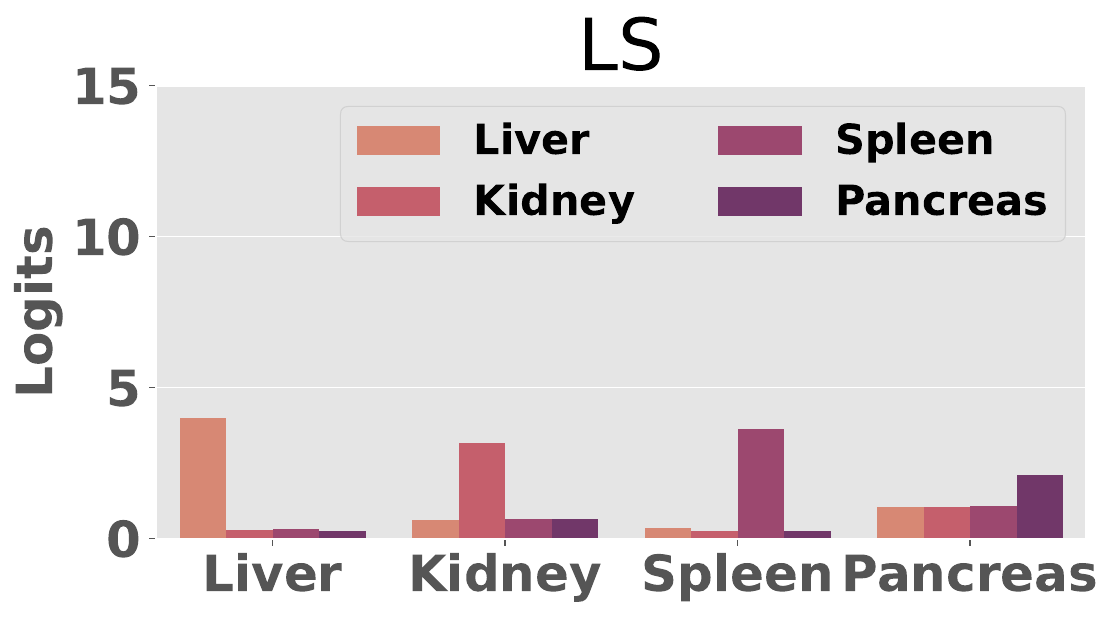}
     \end{subfigure}
     \begin{subfigure}[b]{0.19\linewidth}
         \centering
         \includegraphics[width=\linewidth]{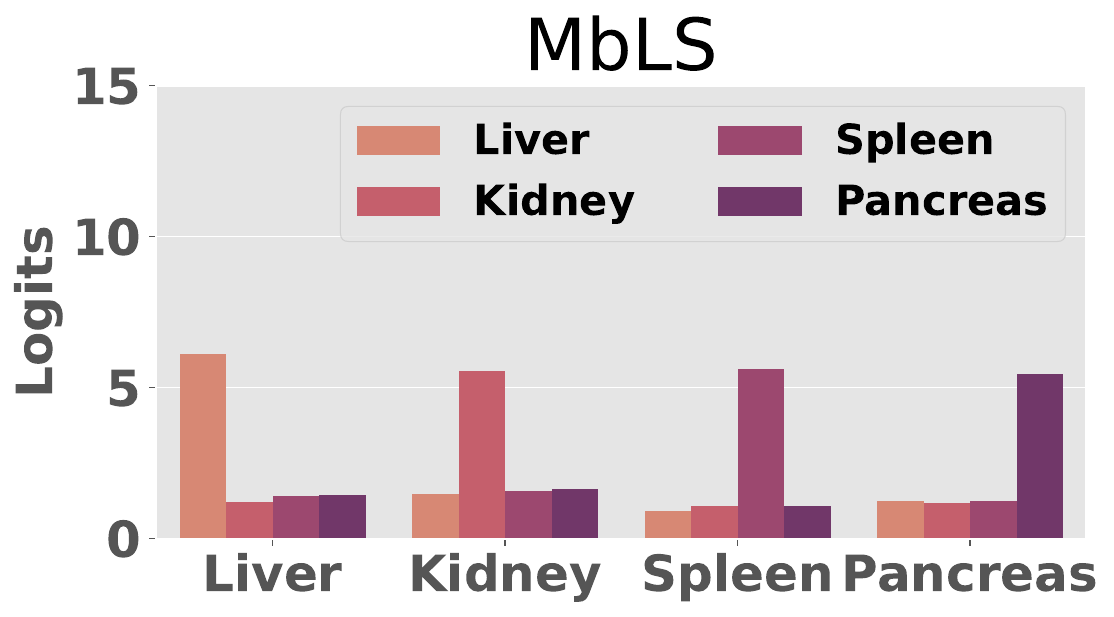}
     \end{subfigure}
     \begin{subfigure}[b]{0.19\linewidth}
         \centering
         \includegraphics[width=\linewidth]{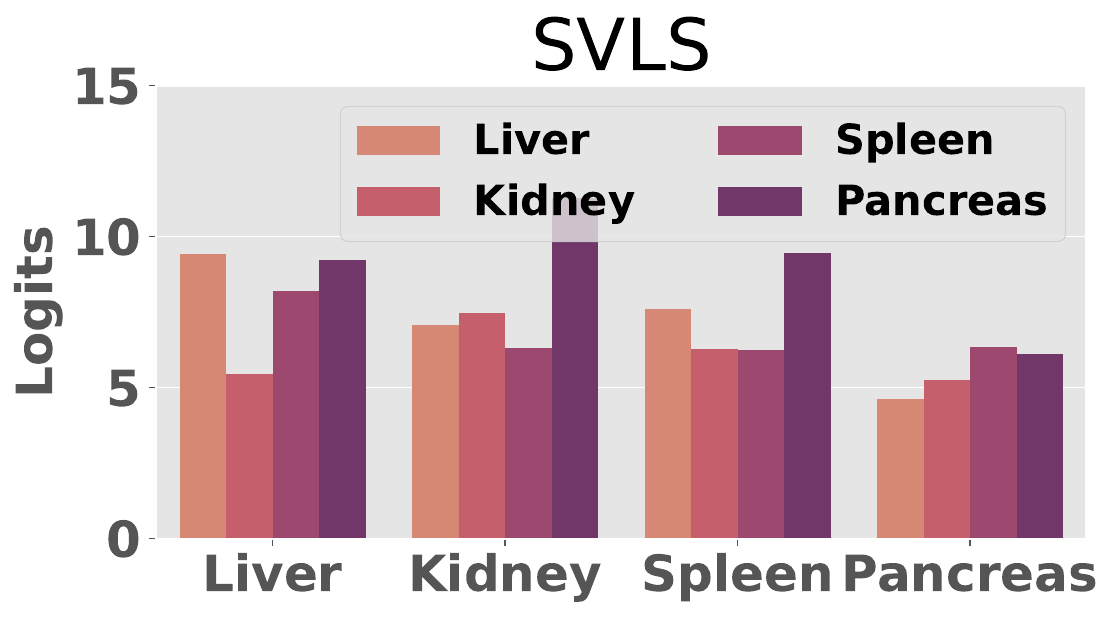}
     \end{subfigure}
     \begin{subfigure}[b]{0.19\linewidth}
         \centering
         \includegraphics[width=\linewidth]{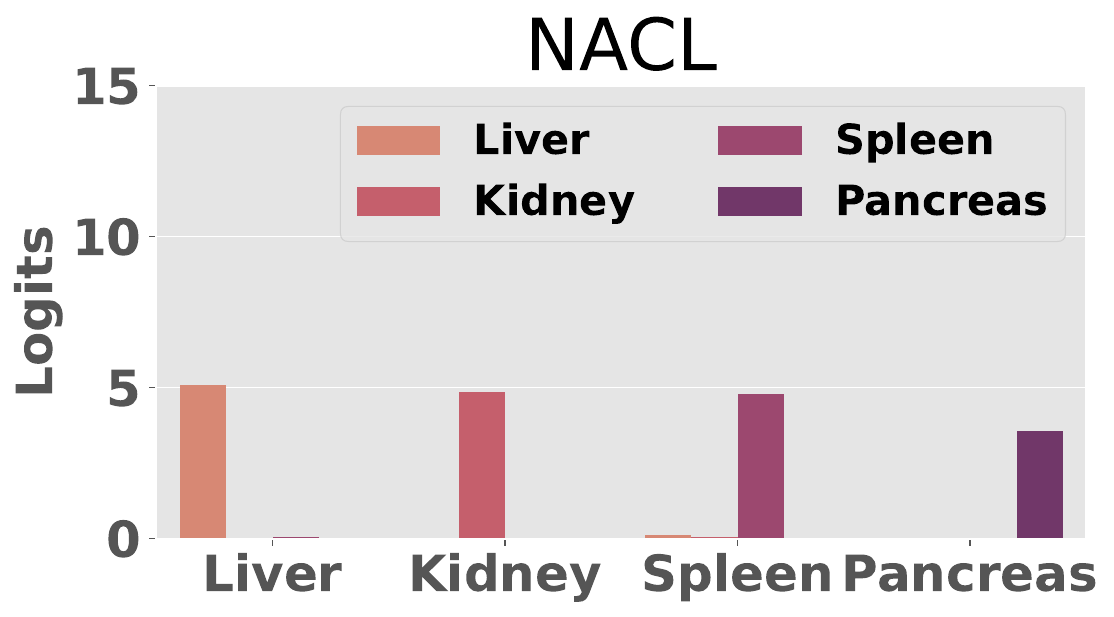}
     \end{subfigure}
     \begin{subfigure}[b]{0.19\linewidth}
         \centering
         \includegraphics[width=\linewidth]{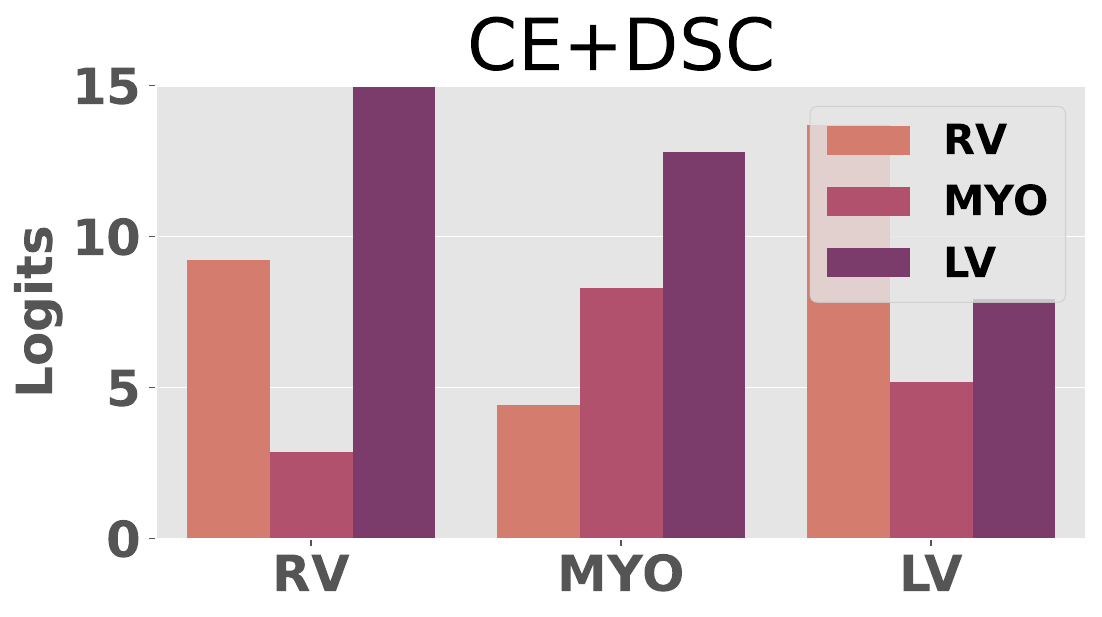}
     \end{subfigure}
     \begin{subfigure}[b]{0.2\linewidth}
         \centering
         \includegraphics[width=\linewidth]{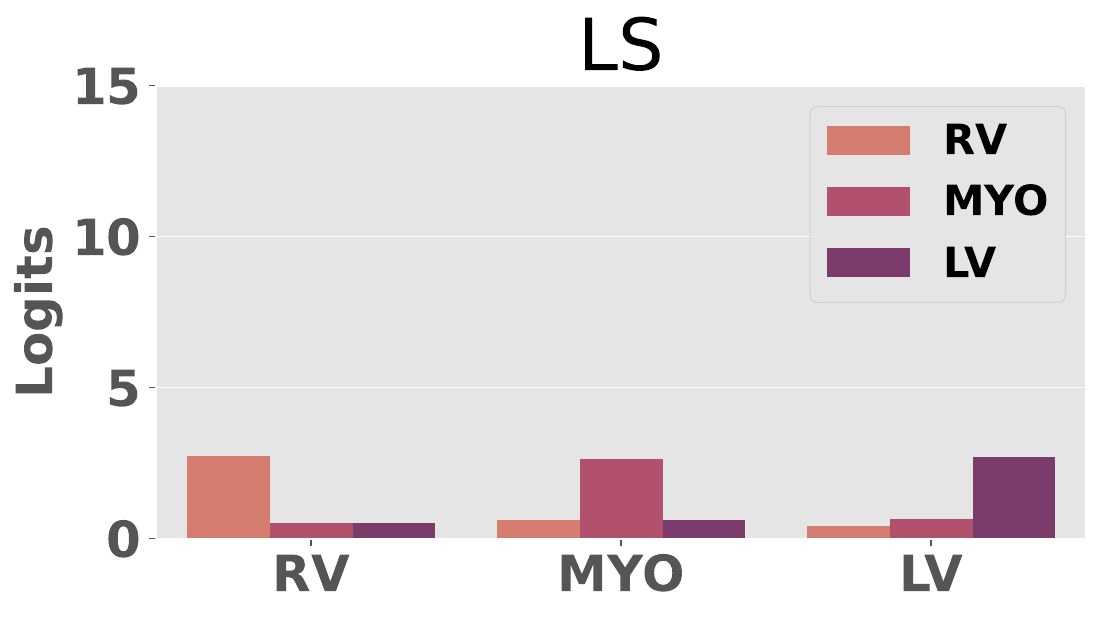}
     \end{subfigure}
     \begin{subfigure}[b]{0.2\linewidth}
         \centering
         \includegraphics[width=\linewidth]{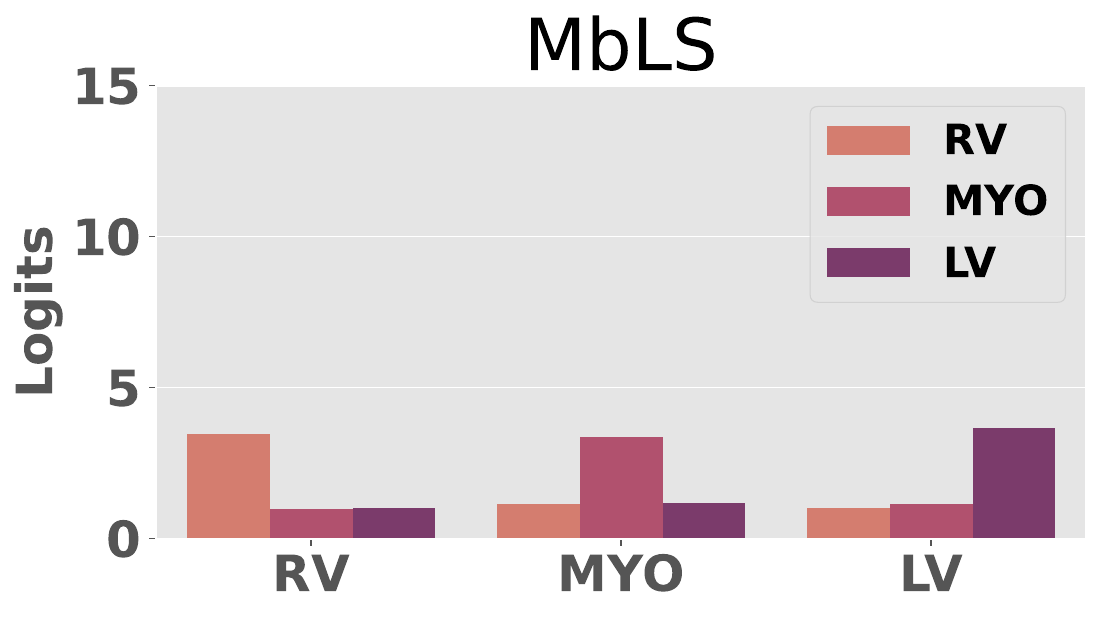}
     \end{subfigure}
     \begin{subfigure}[b]{0.2\linewidth}
         \centering
         \includegraphics[width=\linewidth]{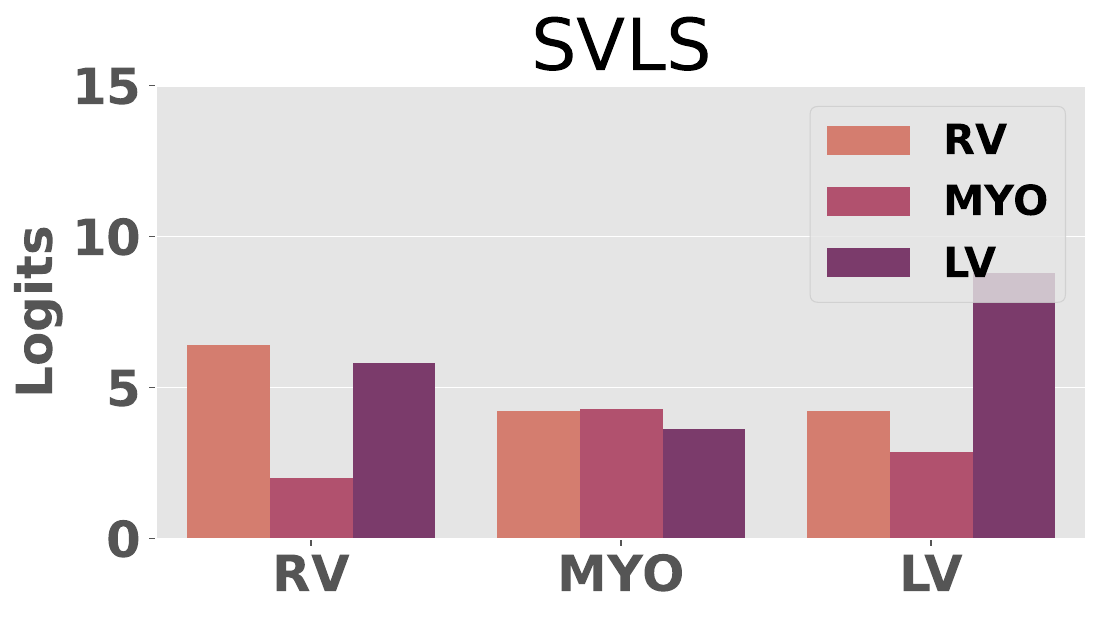}
     \end{subfigure}
     \begin{subfigure}[b]{0.19\linewidth}
         \centering
         \includegraphics[width=\linewidth]{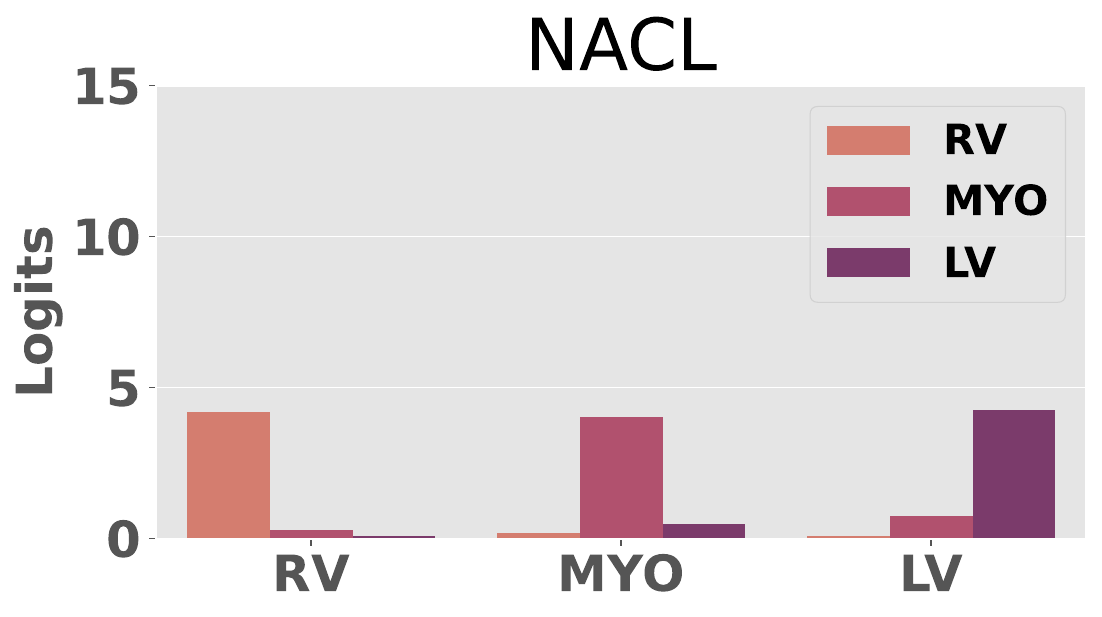}
     \end{subfigure}
    \caption{\rev{Distribution of logit predictions provided by a model trained with CE+DSC, LS, MbLS, SVLS and our approach (\textit{from left to right}) on FLARE (\textit{top}) and  ACDC (\textit{bottom}).}}
    \label{fig:logit-dist}
\end{figure*}

\begin{figure*}[h!]
    \begin{center}
    \includegraphics[width=\linewidth]{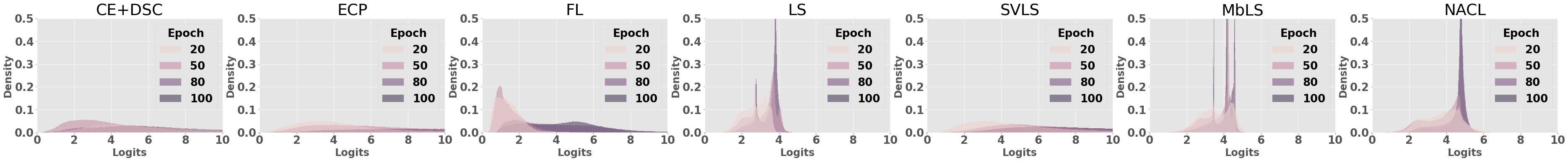}
    \end{center}
    \vspace{-1 em}
    \caption{\rev{Histogram of global logit distribution over epochs obtained by the different approaches. }}
     \label{fig:maximum_logit_histogram_epochs}
\end{figure*}

\noindent \textbf{\rev{Effect on the logit distributions.}}
\rev{In order to demonstrate the benefits of our method over existing approaches, in terms of controlling the logits, we have plotted the average logit distribution across classes on ACDC and FLARE test sets in Figure \ref{fig:logit-dist}. In particular, we first separate all the voxels based on their ground truth labels. Then, for each category, we average the per-voxel vector of logit predictions across each category (in absolute value). From the figure, it can be inferred that the popular CE+DSC loss provides higher logit values for the winner class, and the distance between the winner logits and rest are large, typical characteristics of an overconfident model \citep{murugesan2022calibrating}. 
Interestingly, SVLS seems to follow the logit distribution of CE+DSC, up to a given extent, even though it was designed to emulate LS, but integrating class spatial information. In contrast, whereas LS and MbLS have a desired logit distribution for calibration, particularly for the winner class, the distance with the remaining categories is shorter. This may have an undesirable effect, as predictions where the distance between the winner and remaining logits are very small may lack semantic information needed for maintaining the discriminative performance. Finally, our approach brings the best of both worlds, i.e., it keeps the magnitude of the winner logit low, which facilitates the training of a well-calibrated model, effectively pushes the remaining logit values to a considerable distance, thereby preserving robust discriminative power. }

\rev{To further understand how the different methods control the logit predictions, we plot the maximum logit distribution over epochs during training, which is depicted in Fig. \ref{fig:maximum_logit_histogram_epochs}. It is well known that, calibrated methods show a better regularization, restricting the range of logits to a particular range \citep{muller2019does}. From the figure, it could be observed that, during initial epochs, most of the methods show similar distribution. However, as the number of epochs increases, several methods focusing on improving the calibration performance have a narrower range. Indeed, only LS, MbLS and the proposed NACL approach present the narrowest logit distribution when the network has been trained during a large number of epochs. Based on the findings in \citep{muller2019does}, we can therefore say that our method presents very strong regularization capabilities compared to other approaches, as the range of logits provided by the trained model is very restricted, with most of the logits encountered between a value of 4 and 5.}

\begin{figure*}[h!]
     \centering
     \begin{subfigure}[b]{\linewidth}
         \centering
         \includegraphics[width=\linewidth]{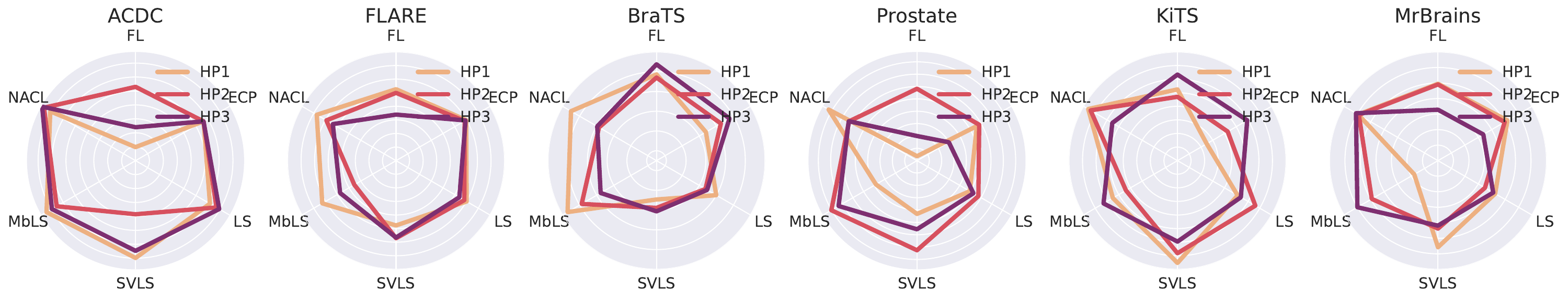}
     \end{subfigure}
     \begin{subfigure}[b]{\linewidth}
         \centering
         \includegraphics[width=\linewidth]{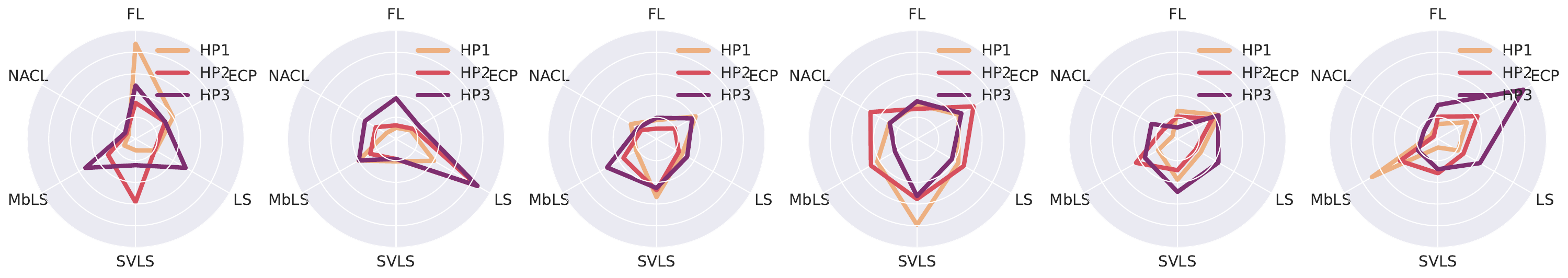}
     \end{subfigure}
    \caption{\rev{\textbf{Radar plots displaying hyperparameter-dependence performance (DSC on \textit{top} and ECE in the \textit{bottom}).} HP1, HP2 and HP3 denote the respective hyper-parameter set: 
    FL ($\gamma$=[1,2,3]), ECP ($\lambda=$[0.1,0.2,0.3]), LS ($\alpha=$[0.1,0.2,0.3]), MbLS ($m$=[3,5,10]) and SVLS ($\sigma$=[0.5,1,2], and ours ($\lambda$=[0.1,0.2,0.3]). Our method consistency provides best performance for $0.1$ across datasets.}}
    \label{fig:radarplot-hp}
\end{figure*}

\subsubsection{\rev{On the impact of hyperparameters}}
\rev{In this experiment, we assess the sensitivity of the hyper-parameters in the different methods, and possibly find a setting which works best across datasets. For FL, $\gamma$ values of 1, 2, and 3 are considered. In the case of ECP and LS, $\alpha$ and $\lambda$ are set to of 0.1, 0.2 and 0.3. For MbLS, we considered the margins to be 5, 8, and 10, while $\lambda$ was fixed to $0.1$. In the case of SVLS, we fixed the kernel size to 3 and used 0.5, 1, and 2 as sigma values. Finally, we fixed $\lambda$ in our method to 0.1, 0.2 and 0.3. We compared the discriminative (DSC) and calibration (ECE) performances using these hyper-parameters across the different datasets and depicted the results in Figure \ref{fig:radarplot-hp}. From this figure, it can be observed that, our method is fairly consistent with a particular hyper-parameter (HP1). Moreover, varying $\lambda$ does not typically result in drastic performance degradation, which demonstrates the robustness of our approach to different hyper-parameter values. In contrast, other methods presented larger performance variations, as discrimination and calibration metrics were highly sensitive to the hyper-parameter choice. For example, in ACDC, focal loss and SVLS suffer large performance degradation for different values of their respective hyper-parameters, whereas in MrBrainS, ECP and MbLS results considerably decrease across different values of $\lambda$ and $m$, respectively.}

\subsubsection{\rev{Effect of the prior}}

\noindent \textbf{Ablation on different priors.} A benefit of the proposed formulation, particularly compared to SVLS \citep{islam2021spatially}, is that diverse priors can be enforced on the logit distributions. Thus, we now assess the impact of different priors, $\bm \tau$ in our formulation, that can distribute the label distribution. The results presented in Table \ref{table:prior} reveal that selecting a suitable prior can further improve the performance of our model.

\begin{table*}[h!]
\centering
\scriptsize
\caption{\textbf{Impact of using different priors}. We compare the discrimative and calibration performance of our approach across the six datasets when using different priors $\bm \tau$ in Equation \ref{eq:proposed-unconstrained}.}
\begin{tabular}{l | cc | cc | cc| cc | cc | cc | cc }
                    \toprule
                    & \multicolumn{2}{c|}{FLARE} & 
                    \multicolumn{2}{c|}{ACDC} & \multicolumn{2}{c|}{BraTS}  & \multicolumn{2}{c|}{\rev{PROSTATE}} & \multicolumn{2}{c|}{\rev{KiTS}} & \multicolumn{2}{c|}{\rev{MRBrainS}} & \multicolumn{2}{c}{\rev{Mean}} \\
                    \midrule
                    Prior $\bm \tau $& DSC & ECE & DSC & ECE & DSC & ECE & DSC & ECE & DSC & ECE & DSC & ECE & DSC & ECE \\
    \midrule
Mean & 0.868 & 0.033 & 0.854 & 0.048 & 0.850 & 0.112 & 0.607 & 0.253 & 0.652 & 0.090 & 0.797 & 0.027 & 0.771 & 0.094 \\
Gaussian & 0.860 & 0.033 & 0.876 & 0.042 & 0.813 & 0.140 & 0.559 & 0.293 & 0.615 & 0.134 & 0.779 & 0.045 & 0.750 & 0.115 \\
\bottomrule
\end{tabular}
\label{table:prior}
\end{table*}

\begin{figure}[h!]
     \centering
     \begin{subfigure}[b]{\linewidth}
         \centering
         \includegraphics[width=\linewidth]{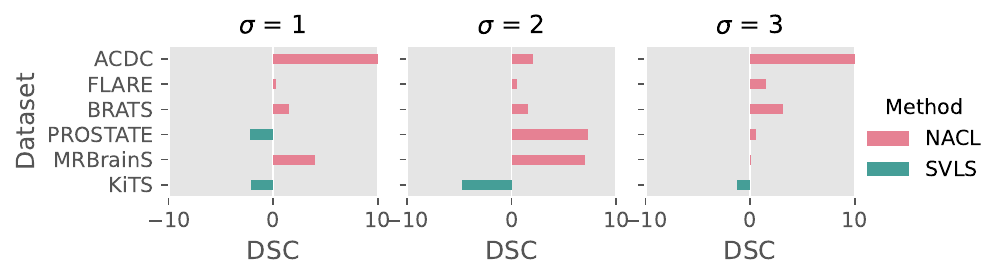}
     \end{subfigure}
     \begin{subfigure}[b]{\linewidth}
         \centering
         \includegraphics[width=\linewidth]{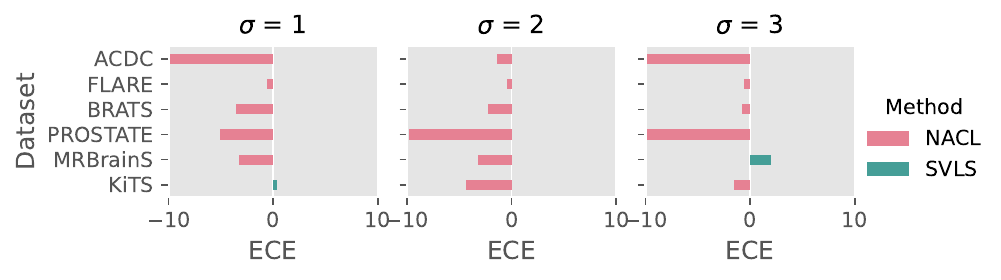}
     \end{subfigure}
    \caption{\rev{\textbf{Direct comparison of SVLS \citep{islam2021spatially} \textit{vs.} NACL (Ours).} Relative error differences (\%) between SVLS and our method when using the same Gaussian prior (with $\sigma=\{1, 2, 3\}$).}}
    \label{fig:gaussian_sigma_study}
\end{figure}

\noindent \textbf{\rev{Varying sigma with a Gaussian prior.}}
\rev{One of the advantages of the proposed approach compared to SVLS is its flexibility to include any prior in the constraint, as well as the integration of a blending parameter that controls the influence of the constraint during training. We now compare the impact of employing different sigma values in both SVLS and our approach. In particular, we use the following values in the Gaussian filter ($\sigma=\{1,2,3\}$) used in SVLS, as well to define a Gaussian prior in our formulation, whose results are depicted in Fig. \ref{fig:gaussian_sigma_study}. 
In this figure, the x-axis represents the relative difference in performance between our method and SVLS. More precisely, if we look at the top row for $\sigma=1$, we can observe that in the ACDC dataset, the proposed approach outperforms SVLS by nearly 10\%, whereas in PROSTATE, SVLS obtains nearly 2\% improvement over our method. Taking this information into account, one can clearly see that, using the same prior, the proposed approach typically outperforms SVLS, and sometimes by a large margin, in both DSC and ECE metrics. 
Importantly, our approach achieves these results even without changing the weighing factor ($\lambda$), as it fixed to $0.1$ to have a fair comparison to SVLS, since SVLS cannot control the importance of the penalty, as exposed in Section \ref{ssec:SVLS}. These results show empirically that our method is able to better leverage the neighboring class information compared to SVLS.} 

\subsubsection{Robustness to backbone}
\begin{figure}[h!]
     \centering
     \begin{subfigure}[b]{0.49\linewidth}
         \centering
         \includegraphics[width=\linewidth]{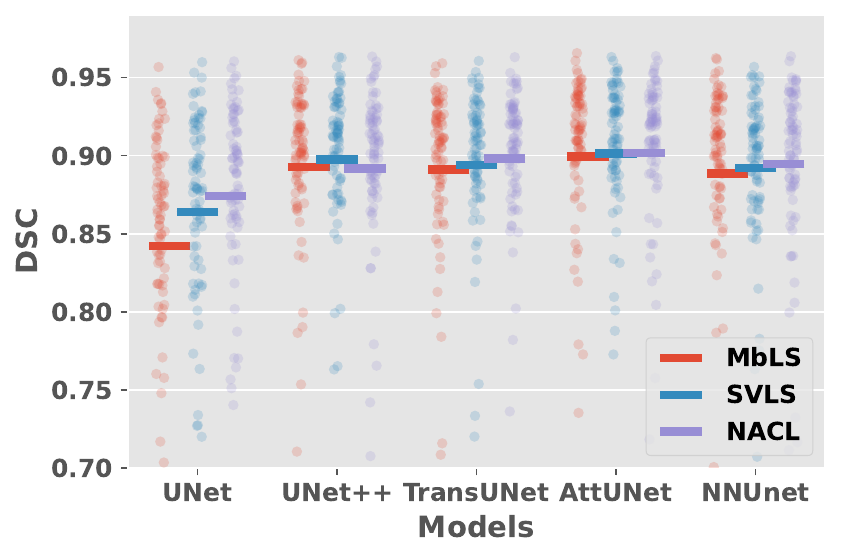}
     \end{subfigure}
     \begin{subfigure}[b]{0.49\linewidth}
         \centering
         \includegraphics[width=\linewidth]{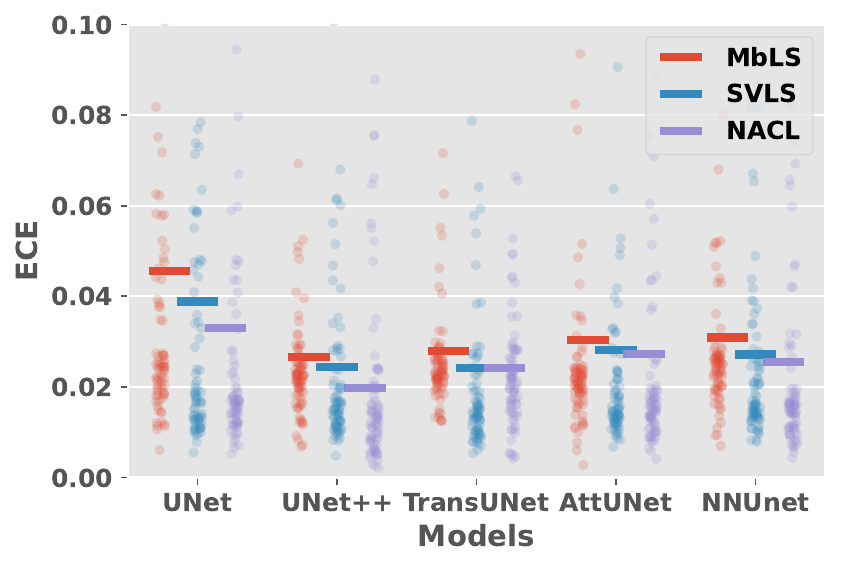}
     \end{subfigure}
    \caption{\rev{\textbf{Robustness to the segmentation backbone.} We evaluate the performance of competing approaches (i.e., MbLS and SVLS) on the FLARE dataset when using different architectures as segmentation backbones.}}
    \label{fig:rob_backbone1}
\end{figure}

\rev{We study the impact of our proposed loss when using other recent state-of-the-art segmentation networks including: AttUNet \citep{attunet}, TransUNet \citep{transunet}, UNet++ \citep{unetpp}, and nnUNet \citep{isensee2021nnu}. We considered the FLARE dataset for this study, whose quantitative results, compared to MbLS and SVLS (our closest competitors in terms of methodology) are presented in Fig. \ref{fig:rob_backbone1}. From the figure, it can be inferred that, regardless of the backbone choice, our method is able to consistently improve both segmentation and calibration performance. This can be attributed to the ability of our method to control the logit distribution, enabling it to be directly plugged into any standard segmentation architecture.}

\subsubsection{Sensitivity to the number of training samples}
\begin{figure}[h!]
     \centering
     \begin{subfigure}[b]{0.49\linewidth}
         \centering
         \includegraphics[width=\linewidth]{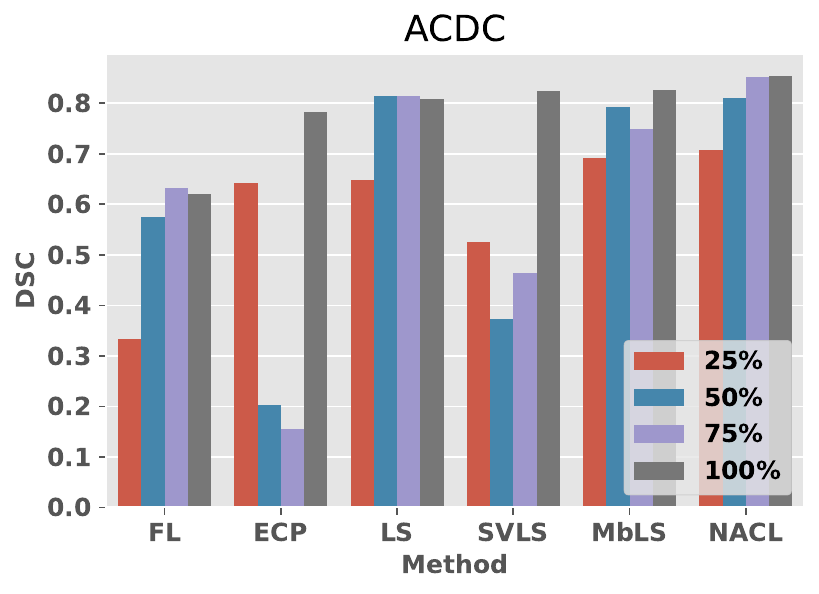}
     \end{subfigure}
     \begin{subfigure}[b]{0.49\linewidth}
         \centering
         \includegraphics[width=\linewidth]{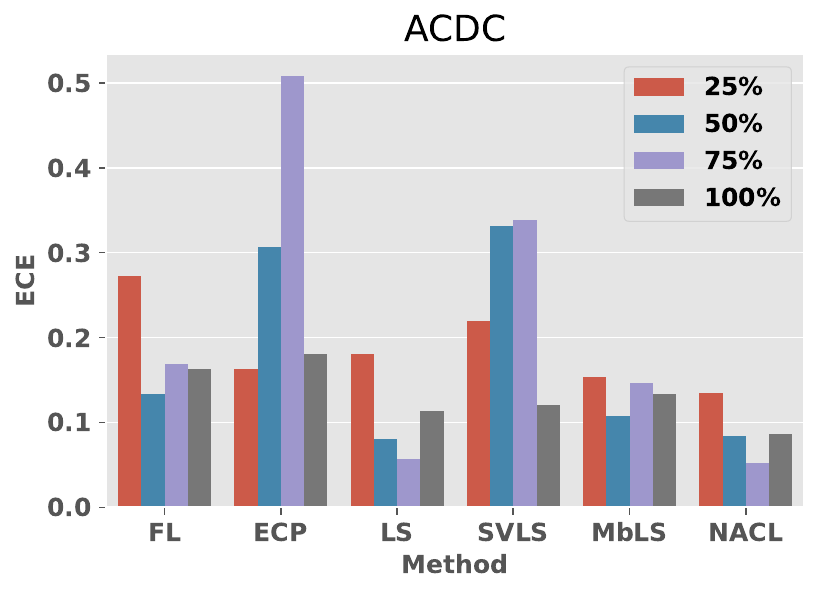}
     \end{subfigure}
     \begin{subfigure}[b]{0.49\linewidth}
         \centering
         \includegraphics[width=\linewidth]{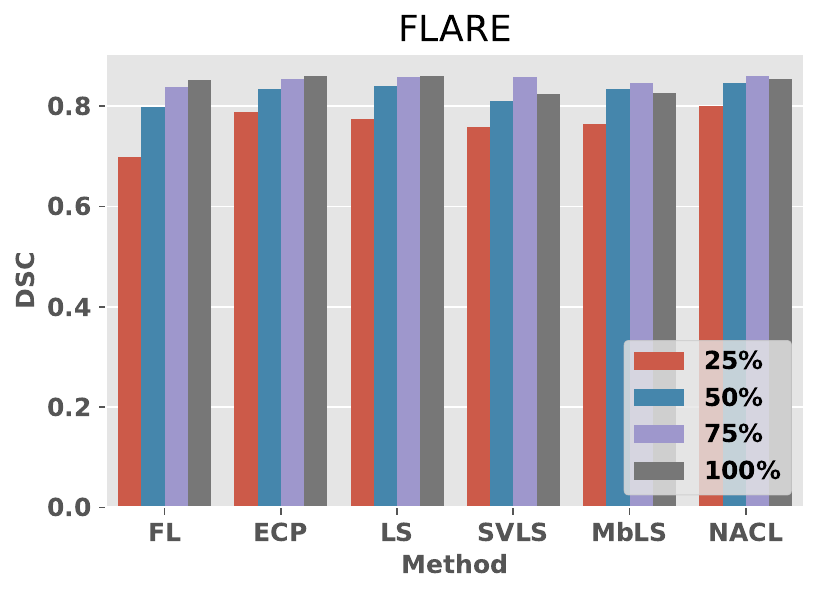}
     \end{subfigure}
     \begin{subfigure}[b]{0.49\linewidth}
         \centering
         \includegraphics[width=\linewidth]{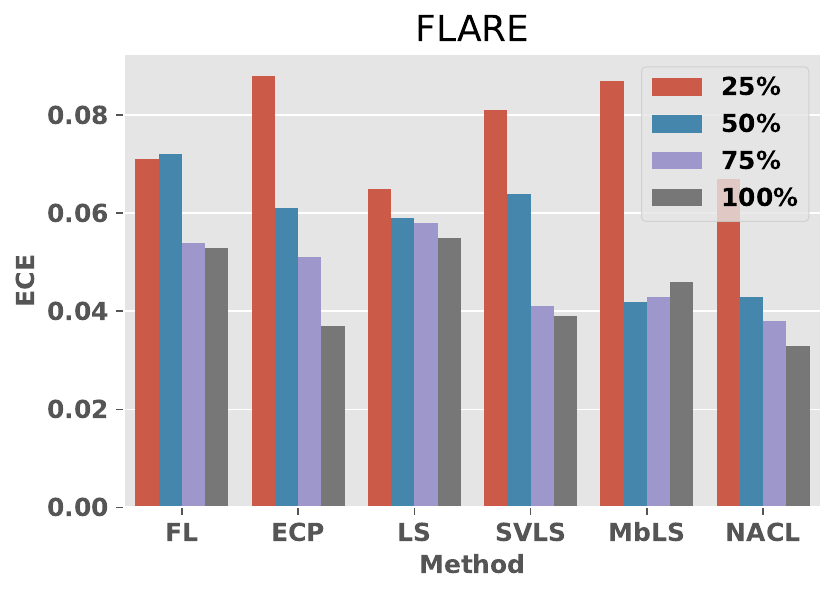}
     \end{subfigure}
    \caption{\rev{\textbf{Performance variation with number of labeled images.} These plots depict the performance of different approaches under several data labeled scenarios, going from 100\% (i.e., original provided data) to 25\% of images from the original dataset.}}
    \label{fig:no_samples}
\end{figure}

\rev{In this experiment, we investigate whether varying the number of training samples impacts the performance of the calibration methods. Indeed, one source of uncertainty in machine learning models is the lack of enough data, which is referred to as \textit{epistemic} uncertainty, or knowledge uncertainty. While this kind of uncertainty can be addressed by adding more knowledge, for example in the form of additional labeled training samples, we want to evaluate how different calibration models behave under different labeled data scenarios. To do so, instead of considering all the samples for training, we only employ 25\%, 50\% and 75\% of the available images. Note that, we use the same validation and test data as we did in our main experiments. Fig. \ref{fig:no_samples} depicts the obtained results for ACDC and FLARE datasets. From these experiments, it is expected that decreasing the number of samples potentially impacts both the discriminative and calibration performance across all the methods. Nevertheless, this trend is not followed by several methods, particularly in terms of correctly modeling the uncertainty. For instance, ECP and SVLS present worst calibration performances for the 50\% and 75\% settings in ACDC, which is also observed in the DSC metrics. Last, across all the labeled scenarios, our approach yields typically the best performance, indicating that it can better handle the epistemic uncertainty derived from lack of enough knowledge during training.}

\subsubsection{Choice of the penalty}
\rev{In this work, we have shown that regularizing the logits based on their neighboring class distribution coupled with the conventional cross entropy is helpful in improving both segmentation and calibration performance. For all the experiments, we have considered a linear penalty to enforce the spatial information. In this section, we now try to control the logits through a quadratic penalty instead. Table \ref{table:l1-l2penalty} presents the comparison of our method with $L_1$ and $L_2$ penalties. From these results, we can observe $L_2$ provides better segmentation results over $L_1$ in more cases, even though in some cases the improvement gains are marginal. Nevertheless, when it underperforms its linear counterpart, the performance gap is significant (e.g., -6\% in PROSTATE). In terms of calibration, $L_1$ yields the best performance in multiple cases. This could be due the nature of $L_2$, which is more aggressive in forcing the logits to follow the prior class distribution compared to $L_1$. It is important to note that, increasing the weighing factor ($\lambda$) of the penalty could mitigate the aggressiveness of $L_2$ to enforce the constraint, potentially leading to the improvement of the segmentation and calibration quality over $L_1$. However, the goal of this work is to provide a unique solution that generalizes across multiple diverse datasets, and that does not require fine-tuning multiple hyper-parameters in each scenario. Thus, we did not explore individual configurations that lead to the best performance for each dataset.}

\begin{table}[h!]
\centering
\footnotesize
\begin{tabular}{l|cc|cc}
\toprule
& \multicolumn{2}{c|}{DSC}  & \multicolumn{2}{c}{ECE} \\
\midrule
& $L_1$ & $L_2$ & $L_1$ & $L_2$ \\
\midrule
ACDC & 0.854 & \textbf{0.871} & \textbf{0.048} & 0.059 \\
FLARE & \textbf{0.868} & 0.851 & \textbf{0.033} & 0.065 \\
BraTS & 0.850 & \textbf{0.851} & 0.112 & \textbf{0.078} \\
\rev{PROSTATE} & \rev{\textbf{0.607}} & \rev{0.541} & \rev{\textbf{0.253}} & \rev{0.320} \\
\rev{KiTS} & \rev{0.652} & \rev{\textbf{0.673}} & \rev{\textbf{0.090}} & \rev{0.106} \\
\rev{MRBrainS} & \rev{0.797} & \rev{\textbf{0.803}} & \rev{0.027} & \rev{\textbf{0.023}} \\ 
\midrule
\rev{Mean} & \rev{\textbf{0.771}} & \rev{0.765} & \rev{\textbf{0.094}} & \rev{0.109} \\ \bottomrule
\end{tabular}
\caption{\textbf{\rev{Impact of different penalties.}} \rev{Comparison of using a $L_1$ \textit{vs} a $L_2$ penalty to impose the constraint in Equation \ref{eq:proposed-unconstrained}.}}
\label{table:l1-l2penalty}
\end{table}

\begin{figure}[h!]
     \centering
     \begin{subfigure}[b]{\linewidth}
         \centering
         \includegraphics[width=0.9\linewidth]{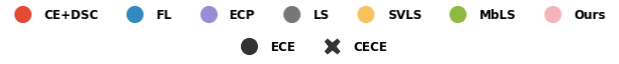}
     \end{subfigure}
     \begin{subfigure}[b]{0.49\linewidth}
         \centering
         \includegraphics[width=\linewidth]{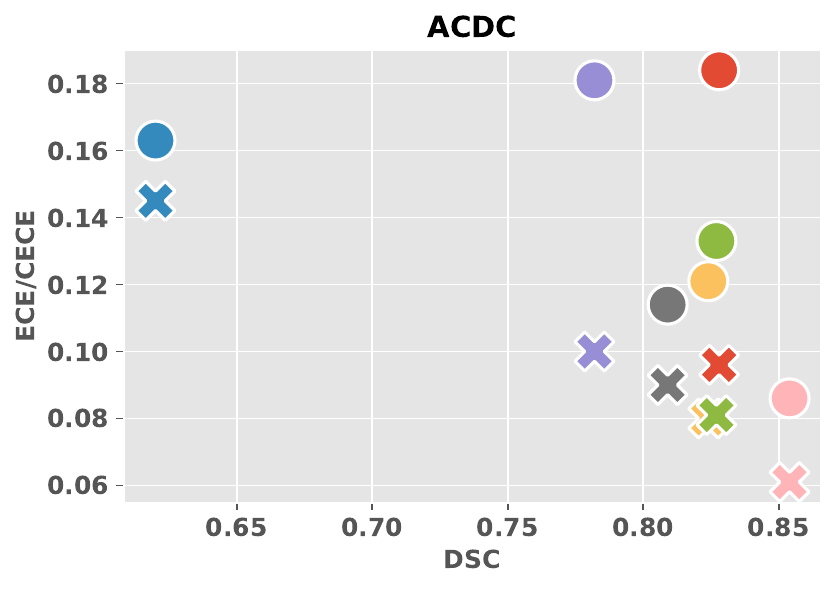}
     \end{subfigure}
     \begin{subfigure}[b]{0.49\linewidth}
         \centering
         \includegraphics[width=\linewidth]{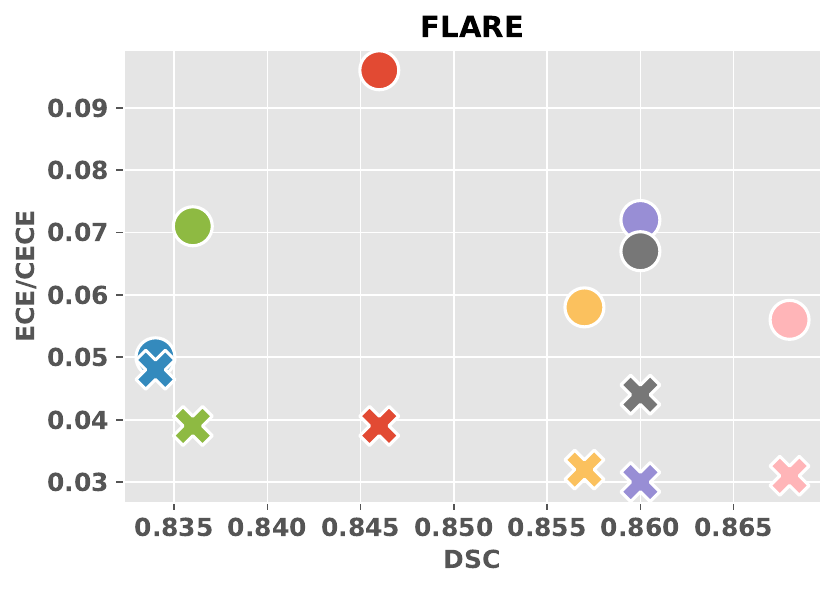}
     \end{subfigure}
     \begin{subfigure}[b]{0.49\linewidth}
         \centering
         \includegraphics[width=\linewidth]{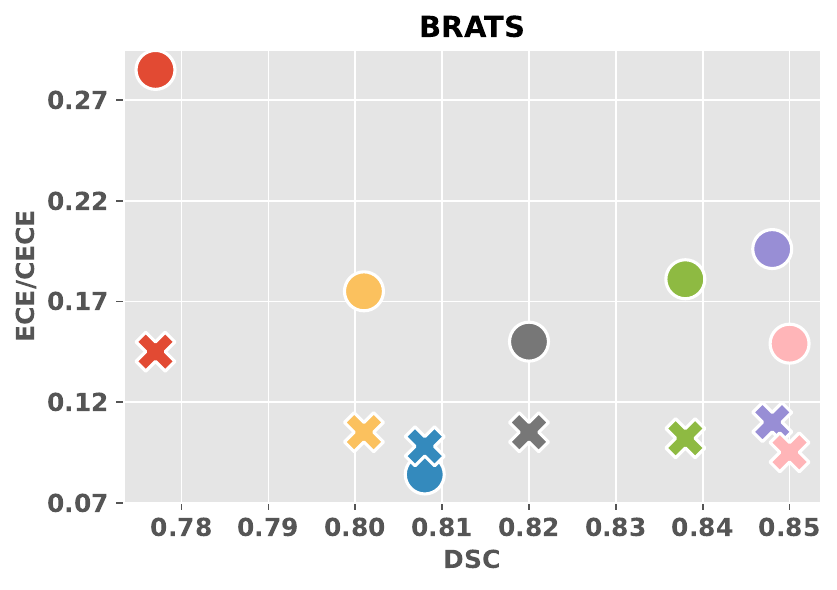}
     \end{subfigure}
     \begin{subfigure}[b]{0.49\linewidth}
         \centering
         \includegraphics[width=\linewidth]{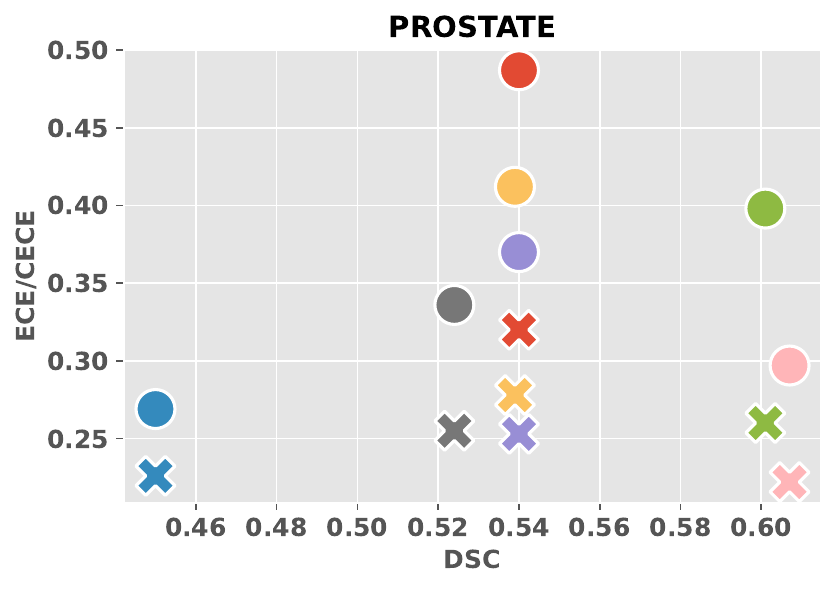}
     \end{subfigure}
     \begin{subfigure}[b]{0.49\linewidth}
         \centering
         \includegraphics[width=\linewidth]{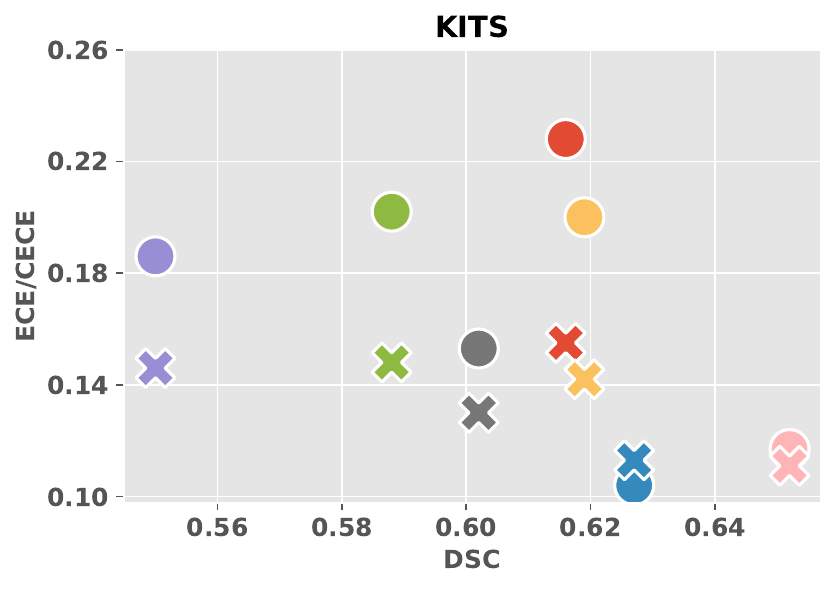}
     \end{subfigure}
     \begin{subfigure}[b]{0.49\linewidth}
         \centering
         \includegraphics[width=\linewidth]{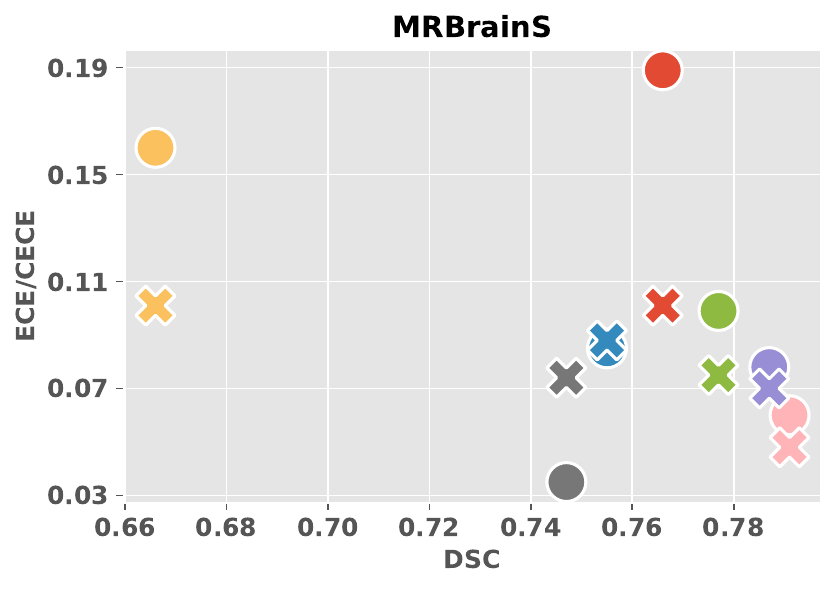}
     \end{subfigure}
    \caption{\rev{Scatter plots comparing DSC vs ECE/CECE when considering the foreground (prediction $\cup$ target) to compute the calibration metrics. 
    }}
    \label{fig:fg-gt-scatter-error}
\end{figure}

\subsubsection{Calibration metrics over prediction and target foregrounds}
\rev{Through all the experiments, the calibration metrics have been obtained by using only the foreground regions of the ground truth. Nevertheless, there is a possibility that a model prediction may be discarded, as it might not overlap with the target ground truth due to an over-segmentation. In this experiment, we recompute the calibration metrics over the union of target and predicted foregrounds, whose ECE and CECE values, against the DSC metric, are depicted in Figure \ref{fig:fg-gt-scatter-error}. We can observe that, even after including the prediction regions in obtaining the calibration metrics, our method still yields the best performance trade-off between DSC and both ECE and CECE across all the datasets. Hence, the strategy for assessing the calibration performance does not change the message that the proposed approach offers a better alternative to existing calibration methods.}

\begin{figure*}[h!]
    \begin{center}
    \includegraphics[width=.95\linewidth]{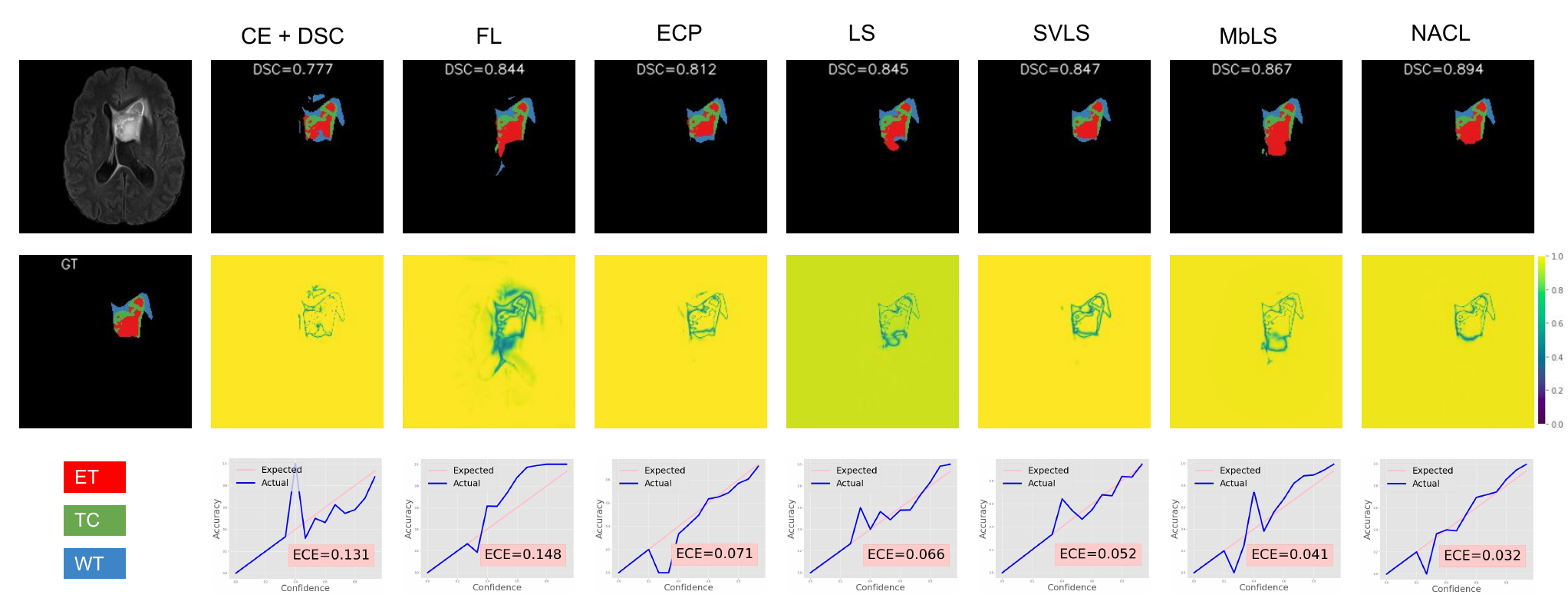}
    \end{center}
    \vspace{-1 em}
    \caption{Qualitative results on BraTS dataset for different methods. In particular, we show the original image and the corresponding segmentation masks provided by each method (\textit{top row}), the ground-truth (GT) mask followed by maximum confidence score of each method (\textit{middle row}) and the respective reliability plots (\textit{bottom row}). Methods from left to right: CE+DSC, FL, ECP, LS, SVLS, MbLS, and Ours}
     \label{fig:qualitative}
\end{figure*}

\subsubsection{Qualitative results and reliability diagrams}
\rev{Last, we show in Figure \ref{fig:qualitative} the predicted segmentation masks (\textit{top}), uncertainty maps (\textit{middle}) and their corresponding reliability plots (\textit{bottom}) on one subject across the different methods. From the predicted segmentation outputs, it is evident that our method generates segmentations closer to the target, which is supported quantitatively by the reported DSC metric. Methods such as MbLS, LS, FL tend to oversegment several categories, whereas ECP and SVLS have difficulties in differentiating challenging regions. The uncertainty maps given by the maximum confidence scores provide more interesting observations on the dynamics of the different methods. Note that, as highlighted in prior works \citep{liu2022devil}, the model should be less confident at the boundaries, while providing more confident predictions in the inner regions. First, we can observe that the CE+DSC compound loss provides the worst calibrated models, as there are no remarkable edges to demarcate between regions. Second, methods such as FL and LS achieve better uncertainty by reducing the overall confidence scores across many regions, which might impact the discriminative performance (as supported by quantitative results reported in previous sections). Third, SVLS provides a distinct edge map, but not particularly sharp because of the smoothing effect of the Gaussian filter. Finally, we could observe that MbLS, as well as our approach, provide confidence estimates that are sharp in the edges and low in within-region pixels, as expected in a well-calibrated model. However, it should be noted that MbLS uses a margin to control the magnitude of the logits, and lacks spatial awareness, as this value is chosen empirically and is equal for all the pixels. This contrasts with our method, where the prior is dynamically chosen depending on the neighboring class distribution for each pixel. Furthermore, we show the our model yields the best reliability diagram, i.e., ECE curves are closer to the diagonal, indicating that the predicted probabilities serve as a good estimate of the correctness of the prediction.}

\section{Conclusion}

\rev{While network calibration has emerged as a mainstay problem in machine learning, most state-of-the-art calibration losses are specifically designed for classification problems, ignoring the spatial information, crucial in dense prediction tasks. Indeed, only the recent SVLS integrates spatial awareness to transform the hard one-hot encoding labels into a smoother version, capturing the class distribution surrounding each pixel. Inspired by the need of leveraging neighboring information to improve the calibration performance of deep segmentation models, in this work we delve into the details of SVLS, and present a constrained optimization perspective of this approach. Our analysis demonstrates that SVLS enforces an implicit constraint on soft class proportions of surrounding pixels. Our formulation exposed two weaknesses of SVLS. First, it lacks a mechanism to
control explicitly the importance of the constraint, which
may hinder the optimization process as it becomes challenging to balance the constraint with the primary objective effectively. And second, the \textit{a priori} knowledge enforced in the constrained is directly derived from the Gaussian distribution of a pixel neighborhood, which may be difficult to define (as it depends on $\sigma$), and did not always provide the best performance, as shown empirically in our results.}

\rev{To overcome the limitations of SVLS, we proposed a principled and simple approach based on equality constraints on the logit values, which allows us to control explicitly both
the prior to be enforced in the constraint, as well as the weight of the penalty, offering more flexibility. We conducted a comprehensive evaluation, incorporating diverse well-known segmentation benchmarks, to evaluate the performance of the proposed approach, and compared it to state-of-the-art calibration losses in the crucial task of medical image segmentation. The empirical findings demonstrate that our approach outperforms existing approaches in both discriminative and calibration metrics. Furthermore, the proposed formulation yields stable results across multiple segmentation backbones, hyper-parameter values, and several labeled data scenarios, establishing itself as a robust alternative within the current literature.}

\rev{While the proposed solution offers superior performance to existing approaches, there exist multiple avenues which are worth to explore. For example, a limitation of our approach is that it disregards image intensity information, which sometimes emerges as the source of annotation uncertainty. Thus, incorporating surrounding image intensity in the constraint could potentially lead to better results. Furthermore, simple penalties (i.e., linear and quadratic) have been explored to enforce the proposed constraint. Integrating more powerful strategies, for example based on log-barrier methods, have shown interesting performance gains in medical imaging problems \citep{kervadec2022constrained}. Therefore, the exploration of these strategies to enforce the imposed constraints could shed light into more powerful alternatives in our formulation. }

\section*{Acknowledgements}
This work is supported by the National Science and Engineering Research Council of Canada (NSERC), via its Discovery Grant program and FRQNT through the Research Support for New Academics program. We also thank Calcul Quebec and Compute Canada.


\bibliographystyle{model2-names-etal.bst}\biboptions{authoryear}
\bibliography{main}

\begin{thebibliography}{41}
\expandafter\ifx\csname natexlab\endcsname\relax\def\natexlab#1{#1}\fi
\providecommand{\url}[1]{\texttt{#1}}
\providecommand{\href}[2]{#2}
\providecommand{\path}[1]{#1}
\providecommand{\DOIprefix}{doi:}
\providecommand{\ArXivprefix}{arXiv:}
\providecommand{\URLprefix}{URL: }
\providecommand{\Pubmedprefix}{pmid:}
\providecommand{\doi}[1]{\href{http://dx.doi.org/#1}{\path{#1}}}
\providecommand{\Pubmed}[1]{\href{pmid:#1}{\path{#1}}}
\providecommand{\bibinfo}[2]{#2}
\ifx\xfnm\relax \def\xfnm[#1]{\unskip,\space#1}\fi
\bibitem[{Antonelli et~al.(2022)Antonelli, Reinke, Bakas, Farahani, Kopp-Schneider, Landman, Litjens, Menze, Ronneberger, Summers et~al.}]{antonelli2022medical}
\bibinfo{author}{Antonelli, M.}, \bibinfo{author}{Reinke, A.}, \bibinfo{author}{Bakas, S.}, et~al., \bibinfo{year}{2022}.
\newblock \bibinfo{title}{{The medical segmentation decathlon}}.
\newblock \bibinfo{journal}{Nature communications} \bibinfo{volume}{13}, \bibinfo{pages}{4128}.
\bibitem[{Bakas et~al.(2017)Bakas, Akbari, Sotiras, Bilello, Rozycki, Kirby, Freymann, Farahani and Davatzikos}]{Bakas2017AdvancingFeaturesJ}
\bibinfo{author}{Bakas, S.}, \bibinfo{author}{Akbari, H.}, \bibinfo{author}{Sotiras, A.}, et~al., \bibinfo{year}{2017}.
\newblock \bibinfo{title}{Advancing the cancer genome atlas glioma mri collections with expert segmentation labels and radiomic features}.
\newblock \bibinfo{journal}{Scientific data} \bibinfo{volume}{4}, \bibinfo{pages}{1--13}.
\bibitem[{Bakas et~al.(2018)Bakas, Reyes, Jakab, Bauer, Rempfler, Crimi, Shinohara, Berger, Ha, Rozycki et~al.}]{Bakas2018IdentifyingChallengeJ}
\bibinfo{author}{Bakas, S.}, \bibinfo{author}{Reyes, M.}, \bibinfo{author}{Jakab, A.}, et~al., \bibinfo{year}{2018}.
\newblock \bibinfo{title}{Identifying the best machine learning algorithms for brain tumor segmentation, progression assessment, and overall survival prediction in the brats challenge}.
\newblock \bibinfo{journal}{arXiv preprint arXiv:1811.02629} .
\bibitem[{Bernard et~al.(2018)Bernard, Lalande, Zotti, Cervenansky, Yang, Heng, Cetin, Lekadir, Camara, Ballester et~al.}]{bernard2018deep}
\bibinfo{author}{Bernard, O.}, \bibinfo{author}{Lalande, A.}, \bibinfo{author}{Zotti, C.}, et~al., \bibinfo{year}{2018}.
\newblock \bibinfo{title}{Deep learning techniques for automatic mri cardiac multi-structures segmentation and diagnosis: is the problem solved?}
\newblock \bibinfo{journal}{IEEE transactions on medical imaging} \bibinfo{volume}{37}, \bibinfo{pages}{2514--2525}.
\bibitem[{Chen et~al.(2021)Chen, Lu, Yu, Luo, Adeli, Wang, Lu, Yuille and Zhou}]{transunet}
\bibinfo{author}{Chen, J.}, \bibinfo{author}{Lu, Y.}, \bibinfo{author}{Yu, Q.}, et~al., \bibinfo{year}{2021}.
\newblock \bibinfo{title}{{Transunet: Transformers make strong encoders for medical image segmentation}}.
\newblock \bibinfo{journal}{arXiv preprint arXiv:2102.04306} .
\bibitem[{Ding et~al.(2021)Ding, Han, Liu and Niethammer}]{ding2021local}
\bibinfo{author}{Ding, Z.}, \bibinfo{author}{Han, X.}, \bibinfo{author}{Liu, P.}, and \bibinfo{author}{Niethammer, M.}, \bibinfo{year}{2021}.
\newblock \bibinfo{title}{Local temperature scaling for probability calibration}, in: \bibinfo{booktitle}{Proceedings of the IEEE/CVF International Conference on Computer Vision}, pp. \bibinfo{pages}{6889--6899}.
\bibitem[{Fort et~al.(2019)Fort, Hu and Lakshminarayanan}]{fort2019deep}
\bibinfo{author}{Fort, S.}, \bibinfo{author}{Hu, H.}, and \bibinfo{author}{Lakshminarayanan, B.}, \bibinfo{year}{2019}.
\newblock \bibinfo{title}{{Deep ensembles: A loss landscape perspective}}.
\newblock \bibinfo{journal}{arXiv preprint arXiv:1912.02757} .
\bibitem[{Gal and Ghahramani(2016)}]{gal2016dropout}
\bibinfo{author}{Gal, Y.} and \bibinfo{author}{Ghahramani, Z.}, \bibinfo{year}{2016}.
\newblock \bibinfo{title}{Dropout as a bayesian approximation: Representing model uncertainty in deep learning}, in: \bibinfo{booktitle}{international conference on machine learning}, \bibinfo{organization}{PMLR}. pp. \bibinfo{pages}{1050--1059}.
\bibitem[{Guo et~al.(2017)Guo, Pleiss, Sun and Weinberger}]{guo2017calibration}
\bibinfo{author}{Guo, C.}, \bibinfo{author}{Pleiss, G.}, \bibinfo{author}{Sun, Y.}, and \bibinfo{author}{Weinberger, K.Q.}, \bibinfo{year}{2017}.
\newblock \bibinfo{title}{On calibration of modern neural networks}, in: \bibinfo{booktitle}{International conference on machine learning}, \bibinfo{organization}{PMLR}. pp. \bibinfo{pages}{1321--1330}.
\bibitem[{Heller et~al.(2019)Heller, Sathianathen, Kalapara, Walczak, Moore, Kaluzniak, Rosenberg, Blake, Rengel, Oestreich et~al.}]{heller2019kits19}
\bibinfo{author}{Heller, N.}, \bibinfo{author}{Sathianathen, N.}, \bibinfo{author}{Kalapara, A.}, et~al., \bibinfo{year}{2019}.
\newblock \bibinfo{title}{{The kits19 challenge data: 300 kidney tumor cases with clinical context, ct semantic segmentations, and surgical outcomes}}.
\newblock \bibinfo{journal}{arXiv preprint arXiv:1904.00445} .
\bibitem[{Isensee et~al.(2021)Isensee, Jaeger, Kohl, Petersen and Maier-Hein}]{isensee2021nnu}
\bibinfo{author}{Isensee, F.}, \bibinfo{author}{Jaeger, P.F.}, \bibinfo{author}{Kohl, S.A.}, et~al., \bibinfo{year}{2021}.
\newblock \bibinfo{title}{{nnU-Net: a self-configuring method for deep learning-based biomedical image segmentation}}.
\newblock \bibinfo{journal}{Nature methods} \bibinfo{volume}{18}, \bibinfo{pages}{203--211}.
\bibitem[{Islam and Glocker(2021)}]{islam2021spatially}
\bibinfo{author}{Islam, M.} and \bibinfo{author}{Glocker, B.}, \bibinfo{year}{2021}.
\newblock \bibinfo{title}{Spatially varying label smoothing: Capturing uncertainty from expert annotations}, in: \bibinfo{booktitle}{International Conference on Information Processing in Medical Imaging}, pp. \bibinfo{pages}{677--688}.
\bibitem[{Jena and Awate(2019)}]{jena2019bayesian}
\bibinfo{author}{Jena, R.} and \bibinfo{author}{Awate, S.P.}, \bibinfo{year}{2019}.
\newblock \bibinfo{title}{{A bayesian neural net to segment images with uncertainty estimates and good calibration}}, in: \bibinfo{booktitle}{International Conference on Information Processing in Medical Imaging}, pp. \bibinfo{pages}{3--15}.
\bibitem[{Jungo et~al.(2020)Jungo, Balsiger and Reyes}]{jungo2020analyzing}
\bibinfo{author}{Jungo, A.}, \bibinfo{author}{Balsiger, F.}, and \bibinfo{author}{Reyes, M.}, \bibinfo{year}{2020}.
\newblock \bibinfo{title}{{Analyzing the quality and challenges of uncertainty estimations for brain tumor segmentation}}.
\newblock \bibinfo{journal}{Frontiers in neuroscience} \bibinfo{volume}{14}, \bibinfo{pages}{282}.
\bibitem[{Karimi and Gholipour(2022)}]{karimi2022improving}
\bibinfo{author}{Karimi, D.} and \bibinfo{author}{Gholipour, A.}, \bibinfo{year}{2022}.
\newblock \bibinfo{title}{{Improving Calibration and Out-of-Distribution Detection in Deep Models for Medical Image Segmentation}}.
\newblock \bibinfo{journal}{IEEE Transactions on Artificial Intelligence} .
\bibitem[{Kervadec et~al.(2022)Kervadec, Dolz, Yuan, Desrosiers, Granger and Ayed}]{kervadec2022constrained}
\bibinfo{author}{Kervadec, H.}, \bibinfo{author}{Dolz, J.}, \bibinfo{author}{Yuan, J.}, et~al., \bibinfo{year}{2022}.
\newblock \bibinfo{title}{{Constrained deep networks: Lagrangian optimization via log-barrier extensions}}, in: \bibinfo{booktitle}{2022 30th European Signal Processing Conference (EUSIPCO)}, \bibinfo{organization}{IEEE}. pp. \bibinfo{pages}{962--966}.
\bibitem[{Larrazabal et~al.(2021)Larrazabal, Mart{\'\i}nez, Dolz and Ferrante}]{larrazabal2021orthogonal}
\bibinfo{author}{Larrazabal, A.J.}, \bibinfo{author}{Mart{\'\i}nez, C.}, \bibinfo{author}{Dolz, J.}, and \bibinfo{author}{Ferrante, E.}, \bibinfo{year}{2021}.
\newblock \bibinfo{title}{{Orthogonal ensemble networks for biomedical image segmentation}}, in: \bibinfo{booktitle}{Medical Image Computing and Computer Assisted Intervention--MICCAI 2021}, pp. \bibinfo{pages}{594--603}.
\bibitem[{Lin et~al.(2017)Lin, Goyal, Girshick, He and Doll{\'a}r}]{lin2017focal}
\bibinfo{author}{Lin, T.Y.}, \bibinfo{author}{Goyal, P.}, \bibinfo{author}{Girshick, R.}, et~al., \bibinfo{year}{2017}.
\newblock \bibinfo{title}{Focal loss for dense object detection}, in: \bibinfo{booktitle}{Proceedings of the IEEE international conference on computer vision}, pp. \bibinfo{pages}{2980--2988}.
\bibitem[{Liu et~al.(2022)Liu, Ben~Ayed, Galdran and Dolz}]{liu2022devil}
\bibinfo{author}{Liu, B.}, \bibinfo{author}{Ben~Ayed, I.}, \bibinfo{author}{Galdran, A.}, and \bibinfo{author}{Dolz, J.}, \bibinfo{year}{2022}.
\newblock \bibinfo{title}{The devil is in the margin: Margin-based label smoothing for network calibration}, in: \bibinfo{booktitle}{Proceedings of the IEEE/CVF Conference on Computer Vision and Pattern Recognition}, pp. \bibinfo{pages}{80--88}.
\bibitem[{Liu et~al.(2023)Liu, Rony, Galdran, Dolz and Ben~Ayed}]{liu2023class}
\bibinfo{author}{Liu, B.}, \bibinfo{author}{Rony, J.}, \bibinfo{author}{Galdran, A.}, et~al., \bibinfo{year}{2023}.
\newblock \bibinfo{title}{{Class Adaptive Network Calibration}}, in: \bibinfo{booktitle}{Proceedings of the IEEE/CVF Conference on Computer Vision and Pattern Recognition}, pp. \bibinfo{pages}{16070--16079}.
\bibitem[{Ma et~al.(2021)Ma, Zhang, Gu, Zhu, Ge, Zhang, An, Wang, Wang, Liu et~al.}]{ma2021abdomenct}
\bibinfo{author}{Ma, J.}, \bibinfo{author}{Zhang, Y.}, \bibinfo{author}{Gu, S.}, et~al., \bibinfo{year}{2021}.
\newblock \bibinfo{title}{Abdomenct-1k: Is abdominal organ segmentation a solved problem?}
\newblock \bibinfo{journal}{IEEE Transactions on Pattern Analysis and Machine Intelligence} \bibinfo{volume}{44}, \bibinfo{pages}{6695--6714}.
\bibitem[{Maier et~al.(2017)Maier, Menze, {von der Gablentz}, Häni, Heinrich, Liebrand, Winzeck, Basit, Bentley, Chen, Christiaens, Dutil, Egger, Feng, Glocker, Götz, Haeck, Halme, Havaei, Iftekharuddin, Jodoin, Kamnitsas, Kellner, Korvenoja, Larochelle, Ledig, Lee, Maes, Mahmood, Maier-Hein, McKinley, Muschelli, Pal, Pei, Rangarajan, Reza, Robben, Rueckert, Salli, Suetens, Wang, Wilms, Kirschke, Krämer, Münte, Schramm, Wiest, Handels and Reyes}]{MAIER2017250}
\bibinfo{author}{Maier, O.}, \bibinfo{author}{Menze, B.H.}, \bibinfo{author}{{von der Gablentz}, J.}, et~al., \bibinfo{year}{2017}.
\newblock \bibinfo{title}{Isles 2015 - a public evaluation benchmark for ischemic stroke lesion segmentation from multispectral mri}.
\newblock \bibinfo{journal}{Medical Image Analysis} \bibinfo{volume}{35}, \bibinfo{pages}{250--269}.
\bibitem[{Mehrtash et~al.(2020)Mehrtash, Wells, Tempany, Abolmaesumi and Kapur}]{mehrtash2020confidence}
\bibinfo{author}{Mehrtash, A.}, \bibinfo{author}{Wells, W.M.}, \bibinfo{author}{Tempany, C.M.}, et~al., \bibinfo{year}{2020}.
\newblock \bibinfo{title}{{Confidence calibration and predictive uncertainty estimation for deep medical image segmentation}}.
\newblock \bibinfo{journal}{IEEE transactions on medical imaging} \bibinfo{volume}{39}, \bibinfo{pages}{3868--3878}.
\bibitem[{Mendrik et~al.(2015)Mendrik, Vincken, Kuijf, Breeuwer, Bouvy, De~Bresser, Alansary, De~Bruijne, Carass, El-Baz et~al.}]{mendrik2015mrbrains}
\bibinfo{author}{Mendrik, A.M.}, \bibinfo{author}{Vincken, K.L.}, \bibinfo{author}{Kuijf, H.J.}, et~al., \bibinfo{year}{2015}.
\newblock \bibinfo{title}{Mrbrains challenge: online evaluation framework for brain image segmentation in 3t mri scans}.
\newblock \bibinfo{journal}{Computational intelligence and neuroscience} \bibinfo{volume}{2015}.
\bibitem[{Menze et~al.(2015)Menze, Jakab, Bauer, Kalpathy-Cramer, Farahani, Kirby, Burren, Porz, Slotboom, Wiest, Lanczi, Gerstner, Weber, Arbel, Avants, Ayache, Buendia, Collins, Cordier, Corso, Criminisi, Das, Delingette, Demiralp, Durst, Dojat, Doyle, Festa, Forbes, Geremia, Glocker, Golland, Guo, Hamamci, Iftekharuddin, Jena, John, Konukoglu, Lashkari, Mariz, Meier, Pereira, Precup, Price, Raviv, Reza, Ryan, Sarikaya, Schwartz, Shin, Shotton, Silva, Sousa, Subbanna, Szekely, Taylor, Thomas, Tustison, Unal, Vasseur, Wintermark, Ye, Zhao, Zhao, Zikic, Prastawa, Reyes and Van~Leemput}]{Menze2015TheBRATSJ}
\bibinfo{author}{Menze, B.H.}, \bibinfo{author}{Jakab, A.}, \bibinfo{author}{Bauer, S.}, et~al., \bibinfo{year}{2015}.
\newblock \bibinfo{title}{The multimodal brain tumor image segmentation benchmark (brats)}.
\newblock \bibinfo{journal}{IEEE Transactions on Medical Imaging} \bibinfo{volume}{34}, \bibinfo{pages}{1993--2024}.
\bibitem[{Mukhoti et~al.(2020)Mukhoti, Kulharia, Sanyal, Golodetz, Torr and Dokania}]{mukhoti2020calibrating}
\bibinfo{author}{Mukhoti, J.}, \bibinfo{author}{Kulharia, V.}, \bibinfo{author}{Sanyal, A.}, et~al., \bibinfo{year}{2020}.
\newblock \bibinfo{title}{Calibrating deep neural networks using focal loss}.
\newblock \bibinfo{journal}{Advances in Neural Information Processing Systems} \bibinfo{volume}{33}, \bibinfo{pages}{15288--15299}.
\bibitem[{M{\"u}ller et~al.(2019)M{\"u}ller, Kornblith and Hinton}]{muller2019does}
\bibinfo{author}{M{\"u}ller, R.}, \bibinfo{author}{Kornblith, S.}, and \bibinfo{author}{Hinton, G.E.}, \bibinfo{year}{2019}.
\newblock \bibinfo{title}{When does label smoothing help?}
\newblock \bibinfo{journal}{Advances in neural information processing systems} \bibinfo{volume}{32}.
\bibitem[{Murugesan et~al.(2023a)Murugesan, Adiga~Vasudeva, Liu, Lombaert, Ben~Ayed and Dolz}]{murugesan2023trust}
\bibinfo{author}{Murugesan, B.}, \bibinfo{author}{Adiga~Vasudeva, S.}, \bibinfo{author}{Liu, B.}, et~al., \bibinfo{year}{2023}a.
\newblock \bibinfo{title}{{Trust your neighbours: Penalty-based constraints for model calibration}}, in: \bibinfo{booktitle}{International Conference on Medical Image Computing and Computer-Assisted Intervention}, pp. \bibinfo{pages}{572--581}.
\bibitem[{Murugesan et~al.(2023b)Murugesan, Liu, Galdran, Ayed and Dolz}]{murugesan2022calibrating}
\bibinfo{author}{Murugesan, B.}, \bibinfo{author}{Liu, B.}, \bibinfo{author}{Galdran, A.}, et~al., \bibinfo{year}{2023}b.
\newblock \bibinfo{title}{Calibrating segmentation networks with margin-based label smoothing}.
\newblock \bibinfo{journal}{Medical Image Analysis} \bibinfo{volume}{87}, \bibinfo{pages}{102826}.
\bibitem[{Naeini et~al.(2015)Naeini, Cooper and Hauskrecht}]{naeini2015obtaining}
\bibinfo{author}{Naeini, M.P.}, \bibinfo{author}{Cooper, G.}, and \bibinfo{author}{Hauskrecht, M.}, \bibinfo{year}{2015}.
\newblock \bibinfo{title}{Obtaining well calibrated probabilities using bayesian binning}, in: \bibinfo{booktitle}{Twenty-Ninth AAAI Conference on Artificial Intelligence}.
\bibitem[{Niculescu-Mizil and Caruana(2005)}]{niculescu2005predicting}
\bibinfo{author}{Niculescu-Mizil, A.} and \bibinfo{author}{Caruana, R.}, \bibinfo{year}{2005}.
\newblock \bibinfo{title}{{Predicting good probabilities with supervised learning}}, in: \bibinfo{booktitle}{Proceedings of the 22nd international conference on Machine learning}, pp. \bibinfo{pages}{625--632}.
\bibitem[{Oktay et~al.(2018)Oktay, Schlemper, Folgoc, Lee, Heinrich, Misawa, Mori, McDonagh, Hammerla, Kainz, Glocker and Rueckert}]{attunet}
\bibinfo{author}{Oktay, O.}, \bibinfo{author}{Schlemper, J.}, \bibinfo{author}{Folgoc, L.L.}, et~al., \bibinfo{year}{2018}.
\newblock \bibinfo{title}{Attention u-net: Learning where to look for the pancreas}, in: \bibinfo{booktitle}{Medical Imaging with Deep Learning}.
\bibitem[{Ovadia et~al.(2019)Ovadia, Fertig, Ren, Nado, Sculley, Nowozin, Dillon, Lakshminarayanan and Snoek}]{ovadia2019can}
\bibinfo{author}{Ovadia, Y.}, \bibinfo{author}{Fertig, E.}, \bibinfo{author}{Ren, J.}, et~al., \bibinfo{year}{2019}.
\newblock \bibinfo{title}{Can you trust your model's uncertainty? evaluating predictive uncertainty under dataset shift}.
\newblock \bibinfo{journal}{Advances in neural information processing systems} \bibinfo{volume}{32}.
\bibitem[{Pereyra et~al.(2017)Pereyra, Tucker, Chorowski, Kaiser and Hinton}]{pereyra2017regularizing}
\bibinfo{author}{Pereyra, G.}, \bibinfo{author}{Tucker, G.}, \bibinfo{author}{Chorowski, J.}, et~al., \bibinfo{year}{2017}.
\newblock \bibinfo{title}{Regularizing neural networks by penalizing confident output distributions}, in: \bibinfo{booktitle}{International Conference on Learning Representations (ICLR)}.
\bibitem[{Platt et~al.(1999)}]{platt1999probabilistic}
\bibinfo{author}{Platt, J.} et~al., \bibinfo{year}{1999}.
\newblock \bibinfo{title}{{Probabilistic outputs for support vector machines and comparisons to regularized likelihood methods}}.
\newblock \bibinfo{journal}{Advances in large margin classifiers} \bibinfo{volume}{10}, \bibinfo{pages}{61--74}.
\bibitem[{Ronneberger et~al.(2015)Ronneberger, Fischer and Brox}]{unet}
\bibinfo{author}{Ronneberger, O.}, \bibinfo{author}{Fischer, P.}, and \bibinfo{author}{Brox, T.}, \bibinfo{year}{2015}.
\newblock \bibinfo{title}{{U-Net: Convolutional Networks for Biomedical Image Segmentation}}, in: \bibinfo{booktitle}{Medical Image Computing and Computer-Assisted Intervention -- MICCAI 2015}, pp. \bibinfo{pages}{234--241}.
\bibitem[{Szegedy et~al.(2016)Szegedy, Vanhoucke, Ioffe, Shlens and Wojna}]{szegedy2016rethinking}
\bibinfo{author}{Szegedy, C.}, \bibinfo{author}{Vanhoucke, V.}, \bibinfo{author}{Ioffe, S.}, et~al., \bibinfo{year}{2016}.
\newblock \bibinfo{title}{Rethinking the inception architecture for computer vision}, in: \bibinfo{booktitle}{Proceedings of the IEEE conference on computer vision and pattern recognition}, pp. \bibinfo{pages}{2818--2826}.
\bibitem[{Tomani et~al.(2021)Tomani, Gruber, Erdem, Cremers and Buettner}]{tomani2021post}
\bibinfo{author}{Tomani, C.}, \bibinfo{author}{Gruber, S.}, \bibinfo{author}{Erdem, M.E.}, et~al., \bibinfo{year}{2021}.
\newblock \bibinfo{title}{Post-hoc uncertainty calibration for domain drift scenarios}, in: \bibinfo{booktitle}{Proceedings of the IEEE/CVF Conference on Computer Vision and Pattern Recognition}, pp. \bibinfo{pages}{10124--10132}.
\bibitem[{Wang et~al.(2019)Wang, Li, Aertsen, Deprest, Ourselin and Vercauteren}]{wang2019aleatoric}
\bibinfo{author}{Wang, G.}, \bibinfo{author}{Li, W.}, \bibinfo{author}{Aertsen, M.}, et~al., \bibinfo{year}{2019}.
\newblock \bibinfo{title}{{Aleatoric uncertainty estimation with test-time augmentation for medical image segmentation with convolutional neural networks}}.
\newblock \bibinfo{journal}{Neurocomputing} \bibinfo{volume}{338}, \bibinfo{pages}{34--45}.
\bibitem[{Zhang et~al.(2020)Zhang, Kailkhura and Han}]{zhang2020mix}
\bibinfo{author}{Zhang, J.}, \bibinfo{author}{Kailkhura, B.}, and \bibinfo{author}{Han, T.Y.J.}, \bibinfo{year}{2020}.
\newblock \bibinfo{title}{Mix-n-match: Ensemble and compositional methods for uncertainty calibration in deep learning}, in: \bibinfo{booktitle}{International conference on machine learning}, \bibinfo{organization}{PMLR}. pp. \bibinfo{pages}{11117--11128}.
\bibitem[{Zhou et~al.(2020)Zhou, Siddiquee, Tajbakhsh and Liang}]{unetpp}
\bibinfo{author}{Zhou, Z.}, \bibinfo{author}{Siddiquee, M.M.R.}, \bibinfo{author}{Tajbakhsh, N.}, and \bibinfo{author}{Liang, J.}, \bibinfo{year}{2020}.
\newblock \bibinfo{title}{{UNet++: Redesigning Skip Connections to Exploit Multiscale Features in Image Segmentation}}.
\newblock \bibinfo{journal}{IEEE Transactions on Medical Imaging} \bibinfo{volume}{39}, \bibinfo{pages}{1856--1867}.

\end{thebibliography}

\end{document}